\documentclass[11pt]{article}

\usepackage[preprint]{acl}

\usepackage{times}
\usepackage{latexsym}
\usepackage{enumitem}

\usepackage{pifont}

\usepackage[T1]{fontenc}
\usepackage[utf8]{inputenc}

\usepackage{microtype}
\usepackage{tabularx}
\usepackage{inconsolata}

\usepackage{graphicx}
\usepackage{placeins}
\graphicspath{{figures/}}

\usepackage{amsmath,amssymb}

\usepackage{booktabs}
\usepackage{multirow}

\usepackage{algorithm}
\usepackage{algpseudocode}

\usepackage{xcolor}
\definecolor{ForestGreen}{rgb}{0.13, 0.55, 0.13}

\title{AlgoSkill: Learning to Design Algorithms by Scheduling Human-Like Skills}

\author{\textbf{Xinyuan Song}$^{1}$ \quad
    \textbf{Zekun Cai}$^{2,3}$ \quad
    \textbf{Liang Zhao}$^{1}$ \\
    $^{1}$Emory University, Atlanta, GA, USA \quad
    $^{2}$The University of Tokyo, Tokyo, Japan \\
    $^{3}$LocationMind, Tokyo, Japan \\
    \texttt{\{xinyuan.song,liang.zhao\}@emory.edu, caizekun@csis.u-tokyo.ac.jp} \\
}

\date{}

\begin{document}
\maketitle

\begin{abstract}
Designing an algorithm from a natural-language problem statement requires identifying the problem structure, reading constraints, choosing a suitable paradigm, checking correctness, and refining complexity. Existing large language model (LLM) methods often rely on direct generation or generic self-refinement, leaving these steps implicit. We propose \textbf{AlgoSkill}, which models algorithm design as sequential decision-making over a typed library of algorithmic skills, including abstraction, constraint analysis, state design, data-structure selection, proof checking, counterexample construction, and complexity refinement. A learned scheduler proposes skills from the current design state, while a Monte Carlo Tree Search (MCTS) controller explores skill sequences using verification feedback from compilation, testing, stress testing, and complexity analysis. Experiments on competitive programming and combinatorial optimization benchmarks show that AlgoSkill improves over direct LLM generation, chain-of-thought prompting, self-refinement, and MCTS without typed skills. Ablations show that typed skills, verification-based repair, and search-based scheduling each contribute to performance. These results support treating automatic algorithm design as verification-guided skill scheduling rather than one-shot code generation.
Our code is available at: \url{https://github.com/Hik289/algorithm_skill.git}.
\end{abstract}
\section{Introduction}
Designing efficient algorithms from natural-language problem statements is difficult because the solver must do more than write code; this challenge appears across program synthesis, code generation, and competitive-programming benchmarks~\citep{gulwani2011automating,alur2013syntax,hendrycks2021measuring,li2022competition,jain2024livecodebench}. It must understand the input/output format, read the constraints, infer a feasible complexity class, choose an algorithmic paradigm, justify correctness, and refine the solution until it satisfies time and memory limits. Human programmers usually handle this through a sequence of reusable steps: problem abstraction, brute-force construction, state design, data-structure selection, proof checking, and complexity refinement, echoing classic views of algorithm selection and reusable temporal abstractions~\citep{rice1976algorithm,burke2013hyper,sutton1999between}. This suggests that algorithm design is better viewed as a state-dependent process over reasoning skills, rather than as a single generation step.

\begin{figure}[t]
  \centering
  \includegraphics[width=\linewidth]{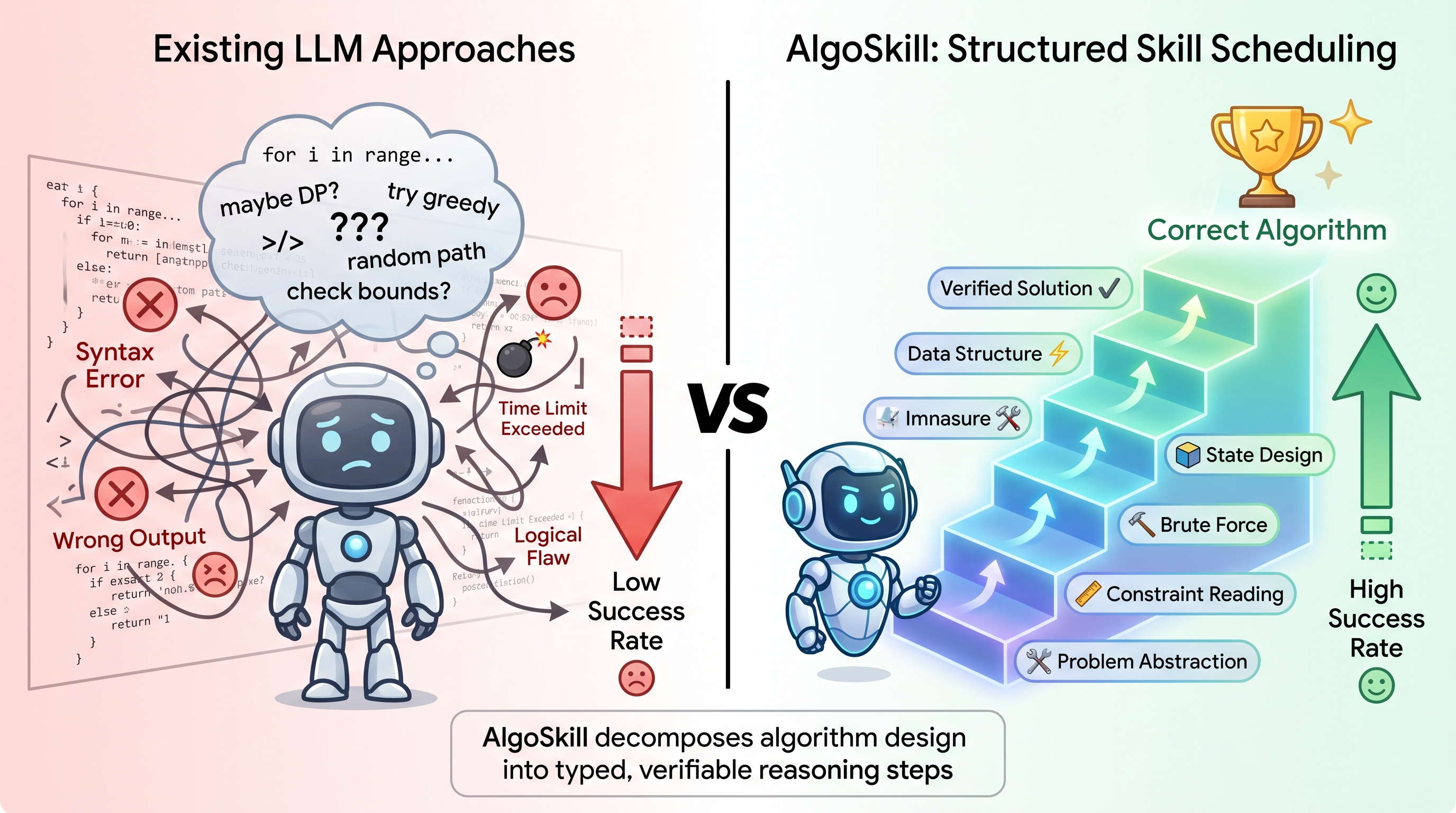}
\caption{\textbf{Core intuition.} Existing LLM methods often rely on free-form generation, which can lead to paradigm, complexity, or proof errors. AlgoSkill structures algorithm design as typed, verifiable skill steps.}
  \label{fig:intuition}
\end{figure}

Large language models (LLMs) have made strong progress in code generation. Codex~\citep{chen2021evaluating}, AlphaCode~\citep{li2022competition}, and DeepSeek-Coder~\citep{guo2024deepseek} perform well on HumanEval and competitive programming benchmarks, especially with large sampling budgets. Prompting and revision methods, such as chain-of-thought reasoning~\citep{wei2022chain}, self-refinement~\citep{madaan2023self}, Reflexion~\citep{shinn2023reflexion}, and Tree of Thoughts~\citep{yao2023tree}, improve multi-step reasoning. Still, these methods usually operate on free-form text or code. They do not make the algorithmic design steps part of the search space. As a result, LLMs can fail on hard problems by choosing the wrong paradigm, missing the intended complexity, defining an incomplete state, or using a greedy rule without a valid argument (Figure~\ref{fig:intuition}).

Search-based systems offer another path. FunSearch~\citep{romera2024mathematical} uses evolutionary program search with LLM-generated mutations, and AlphaEvolve~\citep{novikov2025alphaevolve} extends this idea to broader algorithm design tasks. These systems show that search is useful, but searching over raw programs or free-form solution drafts can be hard to inspect and reuse. A more structured alternative is to search over algorithmic skills, where each step has a typed role and can be checked by verification signals.

Reusable skill libraries have also proved useful in language-agent settings~\citep{chen2023skillit,wang2023voyager}. We propose \textbf{AlgoSkill}, a framework for skill-guided algorithm design. AlgoSkill uses a library of twenty typed algorithmic skills, including problem abstraction, constraint analysis, brute-force construction, state design, data-structure selection, proof checking, counterexample construction, and complexity refinement. Given a design state, a scheduler selects useful skills, and a Monte Carlo Tree Search controller explores skill sequences under feedback from compilation, testing, stress testing, and complexity analysis. In this way, AlgoSkill turns algorithm design into a process of skill selection, composition, and verification.

Our contributions are:
\begin{itemize}[left = 0em]
    \item We define a library of twenty typed algorithmic skills for common steps in algorithm design.
    \item We introduce \textbf{AlgoSkill}, a skill-guided framework that composes these skills under verification feedback.
    \item We evaluate AlgoSkill on competitive programming and combinatorial optimization benchmarks, where it improves over direct generation, prompting, refinement, and search baselines.
\end{itemize}

\begin{figure}[t]
  \centering
  \includegraphics[width=\linewidth]{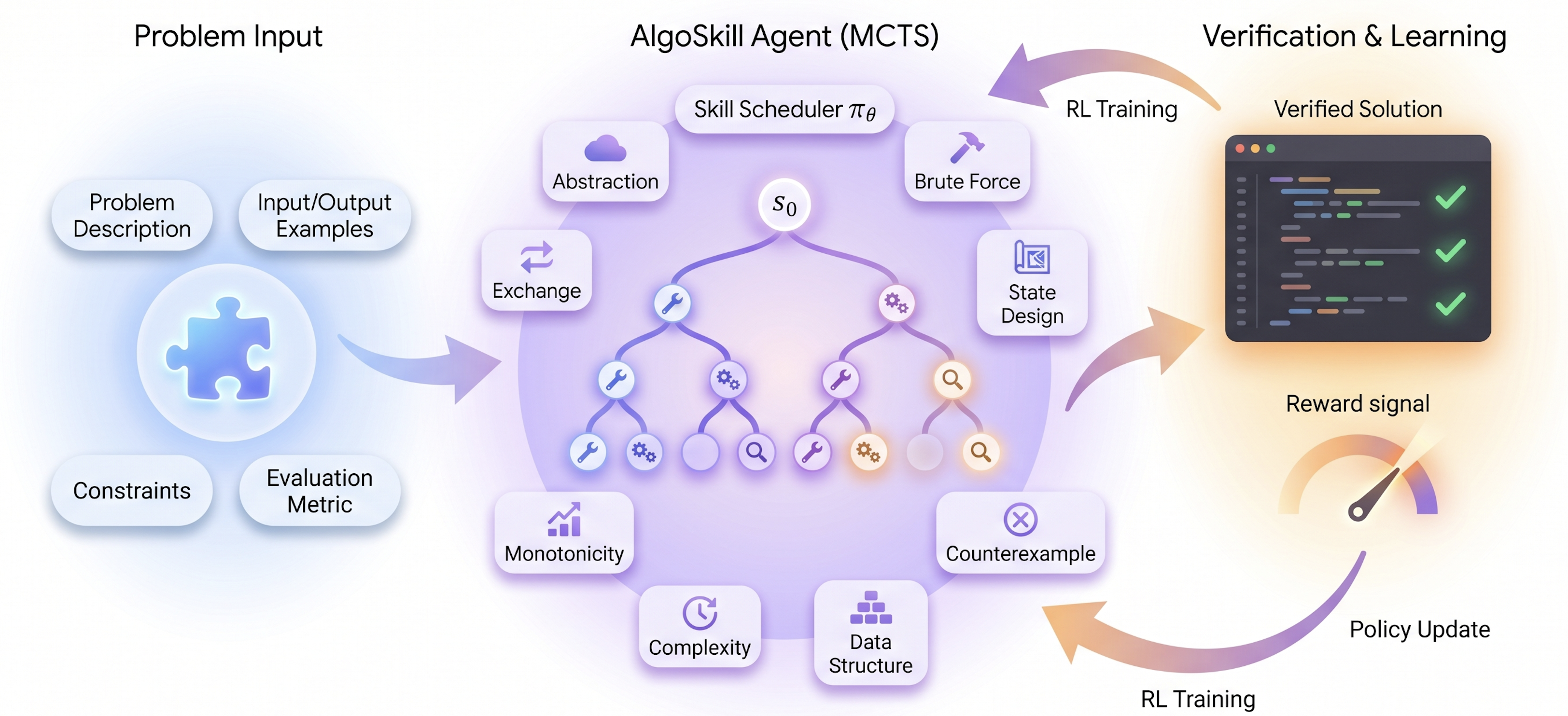}
\caption{\textbf{AlgoSkill overview.} Given a problem $q$, the skill scheduler $\pi_\theta$ uses MCTS to select from twenty typed algorithmic skills. Skill applications update the design state, which is decoded into a candidate algorithm and scored by a verifier to improve scheduling.}
  \label{fig:overview}
\end{figure}

\section{Problem Formulation}
\label{sec:formulation}
As shown in Figure~\ref{fig:overview}, we view algorithm design as a sequence of choices over typed algorithmic skills. A problem instance is written as
\begin{equation}
  q = (\mathcal{D}, \mathcal{I}, \mathcal{O}, \mathcal{K}, \mathcal{M}),
\end{equation}
where $\mathcal{D}$ is the natural-language statement, $\mathcal{I}$ and $\mathcal{O}$ are the input and output formats, $\mathcal{K}$ contains the constraints, and $\mathcal{M}$ is the evaluation metric.

Let $\mathbb{O}=\{o_1,o_2,\ldots,o_N\}$ be the skill library, with $N=20$ in our experiments. At step $t$, AlgoSkill keeps a design state
\begin{equation}
  s_t = (q, r_t, p_t, x_t, \kappa_t, \mathcal{E}_t, h_t) \in \mathcal{S}.
\end{equation}
Here, $r_t$ stores the parsed problem type, task type, and key entities; $p_t$ is the current solution plan; $x_t$ is the current code, if available; $\kappa_t$ is the current time and memory estimate; $\mathcal{E}_t$ stores failed tests, counterexamples, and verifier messages; and
\begin{equation}
  h_t=(o_{0},o_{1},\ldots,o_{t-1})
\end{equation}
records the skills used so far.

At each step, only skills whose preconditions hold can be selected:
\begin{equation}
  \mathbb{O}^{\mathrm{valid}}(s_t)
  =
  \{o_i \in \mathbb{O}: \phi_i^{\mathrm{pre}}(s_t)=1\}.
\end{equation}
The scheduler samples a valid skill
\begin{equation}
  o_t \sim \pi_\theta(\cdot \mid s_t),
  \qquad
  o_t \in \mathbb{O}^{\mathrm{valid}}(s_t),
\end{equation}
and the selected skill updates the state by
\begin{equation}
  s_{t+1}=T(s_t,o_t),
\end{equation}
where $T:\mathcal{S}\times\mathbb{O}\rightarrow\mathcal{S}$ is the skill transition.

After at most $T_{\max}$ steps, the terminal state is decoded into a candidate algorithm,
$a_T=\operatorname{Decode}(s_T)$, including both code and explanation. A full trajectory is
\begin{equation}
  \tau=(s_0,o_0,s_1,o_1,\ldots,s_T,a_T).
\end{equation}
The scheduler is trained to maximize the expected verifier reward:
\begin{equation}
  \max_{\theta}
  \mathbb{E}_{\tau \sim \pi_\theta}
  \left[
    R(a_T,q)
  \right].
\end{equation}
The reward contains correctness, efficiency, explanation, repair, and risk terms:
\begin{equation}
\begin{aligned}
  &R(a_T,q)
  =
  R_{\mathrm{cor}}(a_T,q)
  +
  R_{\mathrm{eff}}(a_T,q)
  \\&+
  R_{\mathrm{expl}}(a_T,q)+
  R_{\mathrm{rep}}(\tau,q)
  -
  R_{\mathrm{risk}}(\tau,q).
  \end{aligned}
\end{equation}
We define these terms in Section~\ref{sec:training}.

\begin{figure*}[t]
  \centering
  \includegraphics[width=0.8\linewidth]{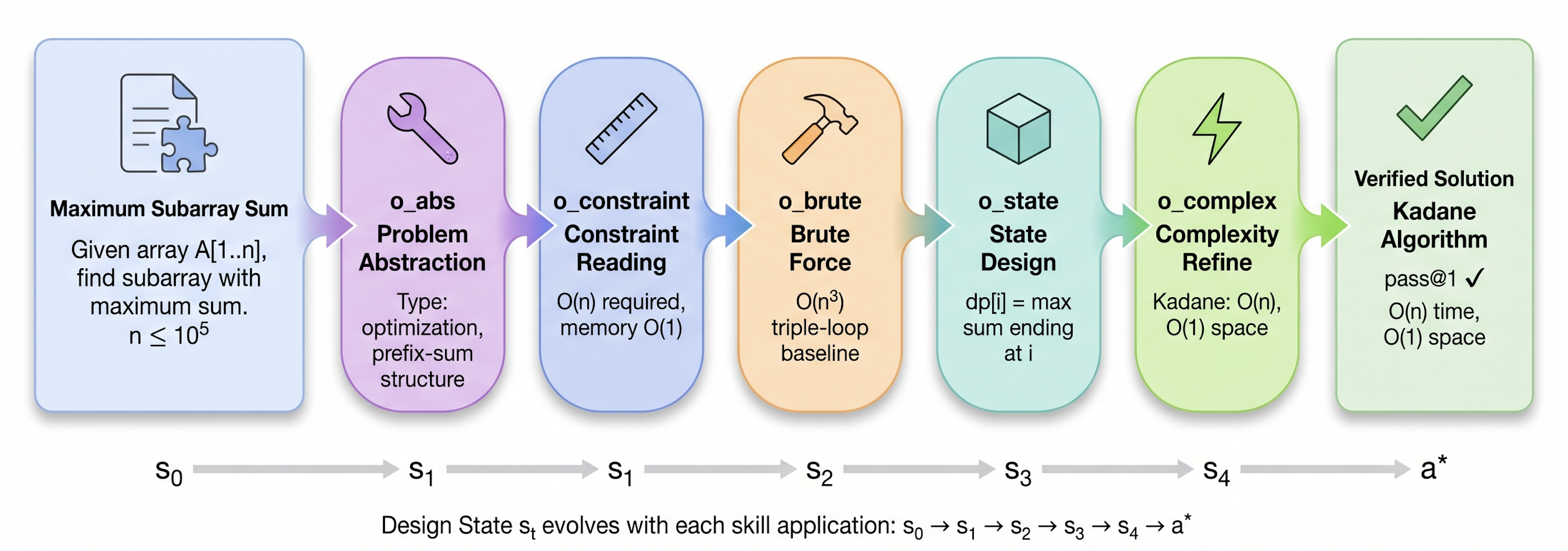}
  \caption{\textbf{Concrete skill trajectory.} Example on \emph{Maximum Subarray Sum}. Starting from $s_0$, AlgoSkill applies $o_{\mathrm{abs}}$ to identify the optimization and prefix-sum structure, $o_{\mathrm{constraint}}$ to infer an $O(n)$ time and $O(1)$ space target, $o_{\mathrm{brute}}$ to build an $O(n^3)$ baseline, $o_{\mathrm{state}}$ to define the DP state $dp[i] = \max$ sum ending at $i$, and $o_{\mathrm{complex}}$ to reach Kadane's $O(n)$ algorithm. Each transition $s_t \to s_{t+1}$ is typed and precondition-checked, and the terminal state decodes to a verified solution.}
  \label{fig:trajectory}
\end{figure*}

This setup keeps the original constraints $\mathcal{K}$ separate from the inferred complexity $\kappa_t$, and separates intermediate code $x_t$ from the final algorithm $a_T$. It also avoids a naming conflict between the output format $\mathcal{O}$ and the skill library $\mathbb{O}$. Compared with direct code generation, AlgoSkill keeps an explicit design state. Compared with free-form self-refinement, each update is a typed skill step. Compared with evolutionary search over programs, it searches over skill trajectories. Figure~\ref{fig:trajectory} gives one example on Maximum Subarray Sum.

\section{Algorithmic Skill Library}
\label{sec:skills}
AlgoSkill uses a typed library of algorithmic skills. Each skill is written as
\begin{equation}
  o_i = (\phi_i^{\mathrm{pre}},\ \tau_i,\ \phi_i^{\mathrm{post}},\ c_i,\ v_i,\ r_i),
\end{equation}
where $\phi_i^{\mathrm{pre}}$ checks whether the skill is applicable, $\tau_i$ updates the design state, $\phi_i^{\mathrm{post}}$ specifies the expected outcome, $c_i$ records the intended complexity effect, $v_i$ defines the verification rule, and $r_i$ lists common failure modes.

The library contains twenty skills:
\begin{align}
  \mathbb{O} = \{&o_{\mathrm{abs}}, o_{\mathrm{constraint}}, o_{\mathrm{brute}}, o_{\mathrm{mono}}, o_{\mathrm{inv}}, o_{\mathrm{state}},\nonumber\\
  &o_{\mathrm{decomp}}, o_{\mathrm{relax}}, o_{\mathrm{exchange}}, o_{\mathrm{ds}}, o_{\mathrm{dual}}, o_{\mathrm{counter}},\nonumber\\
  &o_{\mathrm{complex}}, o_{\mathrm{analogy}}, o_{\mathrm{math}}, o_{\mathrm{sym}}, o_{\mathrm{random}},\nonumber\\
  &o_{\mathrm{approx}}, o_{\mathrm{cache}}, o_{\mathrm{parallel}}\}.
\end{align}

These skills cover the main steps that appear in algorithm design. Early-stage skills, such as $o_{\mathrm{abs}}$ and $o_{\mathrm{constraint}}$, parse the problem and infer the feasible complexity range. Construction skills, such as $o_{\mathrm{brute}}$, $o_{\mathrm{state}}$, and $o_{\mathrm{ds}}$, build candidate algorithms or replace inefficient operations. Verification and repair skills, such as $o_{\mathrm{inv}}$, $o_{\mathrm{exchange}}$, $o_{\mathrm{counter}}$, and $o_{\mathrm{complex}}$, check correctness, test failure cases, and refine inefficient solutions. The remaining skills handle special structures, including decomposition, relaxation, duality, formulas, symmetry, randomization, approximation, caching, and parallelization.

We keep the library typed rather than treating skills as free-form prompts. This lets the scheduler mask invalid actions, track the effect of each skill on the design state, and assign credit to specific algorithmic decisions. The full skill descriptions and the complete library table are given in Appendix~\ref{app:skill_library}.

\section{Skill Scheduling and Search}
\label{sec:search}
AlgoSkill schedules skills in three steps: it first masks invalid skills, then scores the remaining skills with a learned policy, and finally uses MCTS to search over skill sequences.

\subsection{Valid-Skill Filter}

At state $s_t$, a skill can be selected only if its precondition is satisfied:
\begin{equation}
  \mathbb{O}^{\mathrm{valid}}_t
  =
  \{o_i \in \mathbb{O} : \phi_i^{\mathrm{pre}}(s_t)=1\}.
\end{equation}
The preconditions are checked from fields in the design state. For example, if the current complexity estimate suggests a timeout risk, the filter enables $o_{\mathrm{complex}}$, $o_{\mathrm{ds}}$, $o_{\mathrm{decomp}}$, and $o_{\mathrm{mono}}$. If $\mathcal{E}_t$ contains failed tests or verifier messages, the filter enables $o_{\mathrm{counter}}$, $o_{\mathrm{state}}$, $o_{\mathrm{inv}}$, and $o_{\mathrm{exchange}}$. This masking step removes skills that do not fit the current state and reduces the branching factor in search.
\subsection{Learned Skill Policy}
The scheduler uses a neural policy $f_\theta$ to score valid skills:
\begin{equation}
  \pi_\theta(o \mid s_t)
  =
  \operatorname{softmax}(f_\theta(s_t)).
\end{equation}
The input to $f_\theta$ includes the parsed problem type, constraint profile, current plan $p_t$, current complexity estimate $\kappa_t$, recent failure signals in $\mathcal{E}_t$, and skill history $h_t$. A greedy scheduler, used as one baseline, selects
\begin{equation}
  o_t
  =
  \arg\max_{o \in \mathbb{O}^{\mathrm{valid}}_t}
  \pi_\theta(o \mid s_t).
\end{equation}
\subsection{MCTS Search Controller}
The search controller uses MCTS~\citep{silver2017mastering} over design states. Each node is a state $s$, and each edge is a skill application $o$.

\textbf{Selection.}
Starting from $s_0$, the controller selects the skill with the largest UCB score~\citep{auer2002finite}:
\begin{equation}
  o^*
  =
  \operatorname*{argmax}_{o \in \mathbb{O}^{\mathrm{valid}}(s)}
  \left[
    Q(s,o)
    +
    \beta P_\theta(o \mid s)
    \frac{\sqrt{N(s)}}{1+N(s,o)}
  \right].
\end{equation}
Here, $Q(s,o)$ is the current value estimate for applying skill $o$ at state $s$, $P_\theta(o \mid s)$ is the policy prior, $N(s)$ and $N(s,o)$ are visit counts, and $\beta$ controls exploration.

\textbf{Expansion.}
The selected skill produces a new state:
\begin{equation}
  s' = T(s,o^*).
\end{equation}

\textbf{Simulation.}
The new state is decoded into a candidate algorithm,
\begin{equation}
  a = \operatorname{Decode}(s'),
\end{equation}
which is evaluated by the verifier using compilation, unit tests, stress tests, and complexity checks.

\textbf{Backpropagation.}
The verifier reward is propagated to the visited state-skill pairs:
\begin{equation}
  Q(s,o)
  \leftarrow
  (1-\eta)Q(s,o)+\eta R.
\end{equation}
The visit counts are updated by $N(s)\leftarrow N(s)+1$ and $N(s,o)\leftarrow N(s,o)+1$.
\subsection{Symbolic Complexity Tracking}

For complexity checking, AlgoSkill represents an algorithm plan as an operation graph $G_a=(V_a,E_a)$, where nodes are operation blocks and edges denote data flow. The complexity is estimated by simple composition rules:
{\small
\begin{align*}
  C(A;B)          &= C(A)+C(B),
  &
  C(\text{loop}) &= \textstyle\prod_i n_i C(\text{body}),\\
  C(\text{sort}) &= O(n\log n),
  &
  C(\text{BFS})  &= O(|V|+|E|),\\
  C(\text{heap}) &= O(k\log n),
  &
  C(\text{seg})  &= O((u+q)\log n),\\
  C(\text{DP})   &= O(|\mathcal{S}|\cdot|\mathcal{A}|).
\end{align*}
}
Given the maximum input size $n_{\max}$, the time and memory penalties are
\begin{equation}
  P_T(a,q)
  =
  \left[
    \log T(a,n_{\max})-\log T_{\max}
  \right]_+,
\end{equation}
\begin{equation}
  P_M(a,q)
  =
  \left[
    \log M(a,n_{\max})-\log M_{\max}
  \right]_+.
\end{equation}
We also run the implementation on input sizes $n_1<n_2<\cdots<n_k$ and fit
\begin{equation}
  \hat{T}(n)=c n^\alpha(\log n)^\gamma.
\end{equation}
If the fitted exponent $\alpha$ disagrees with the symbolic estimate, AlgoSkill adds a complexity-risk penalty.

Algorithm~\ref{alg:algoskill} gives the full search procedure.

\begin{algorithm}[t]
  \caption{AlgoSkill Search Procedure}
  \label{alg:algoskill}
  \begin{algorithmic}[1]
    \Require Problem $q$, skill library $\mathbb{O}$, scheduler $\pi_\theta$, verifier $V$, search budget $B$
    \Ensure Best verified algorithm $a^*$ found in search tree $\mathcal{T}$
    \State Construct the initial state $s_0$ from $q$
    \State Initialize $\mathcal{T}$ with root $s_0$
    \For{$b=1,\ldots,B$}
      \State Select a state $s$ from $\mathcal{T}$ by tree traversal
      \State Compute valid skills $\mathbb{O}^{\mathrm{valid}}(s)=\{o:\phi_o^{\mathrm{pre}}(s)=1\}$
      \State Select $o^*=\arg\max_o\left[Q(s,o)+\beta P_\theta(o\mid s)\frac{\sqrt{N(s)}}{1+N(s,o)}\right]$
      \State Apply the skill: $s'=T(s,o^*)$
      \State Decode $s'$ into a candidate algorithm $a'$
      \State Estimate symbolic time $T(a')$ and memory $M(a')$
      \State Evaluate $a'$ with verifier $V$ and obtain reward $R(a',q)$
      \State Add $s'$ to $\mathcal{T}$
      \State Update $Q(s,o)\leftarrow(1-\eta)Q(s,o)+\eta R$
      \State Update $N(s)\leftarrow N(s)+1$ and $N(s,o)\leftarrow N(s,o)+1$
      \State Assign local skill credit $\Delta_t=R(s')-R(s)$ and update $Q_\omega(s,o)$
      \State Update $\pi_\theta$ using the trajectory reward
    \EndFor
    \State \Return best verified algorithm $a^*$ found in $\mathcal{T}$
  \end{algorithmic}
\end{algorithm}

\section{Training with Verification Rewards}
\label{sec:training}
AlgoSkill trains the scheduler $\pi_\theta$ with policy-gradient reinforcement learning. The reward is computed by an external verifier, so the scheduler is trained from signals that are tied to execution, complexity, and repair quality rather than from final pass/fail labels alone.
\subsection{Reward Decomposition}
For a trajectory $\tau=(s_0,o_0,\ldots,s_T,a_T)$, the reward is
\begin{align}
  R(a_T,q)
  =
  R_{\mathrm{cor}}
  +
  R_{\mathrm{eff}}
  +
  R_{\mathrm{expl}}
  +
  R_{\mathrm{rep}}
  -
  R_{\mathrm{risk}}.
\end{align}

The correctness term combines several test signals:
\begin{equation}
  R_{\mathrm{cor}}
  =
  w_1 R_{\mathrm{cpl}}
  +
  w_2 R_{\mathrm{unit}}
  +
  w_3 R_{\mathrm{str}}
  +
  w_4 R_{\mathrm{ora}},
\end{equation}
where $R_{\mathrm{cpl}}\in\{0,1\}$ is compilation success, $R_{\mathrm{unit}}\in[0,1]$ is the public-test pass rate, $R_{\mathrm{str}}$ is the stress-test pass rate against the brute-force oracle, and $R_{\mathrm{ora}}$ is the oracle agreement rate.

The efficiency term penalizes time, memory, and empirical runtime violations:
\begin{equation}
  R_{\mathrm{eff}}
  =
  -\lambda_T P_T(a,q)
  -
  \lambda_M P_M(a,q)
  -
  \lambda_R P_{\mathrm{rt}}(a,q).
\end{equation}

The explanation term checks whether the final answer contains the main algorithm idea, pseudocode, complexity analysis, and correctness argument:
\begin{equation}
  R_{\mathrm{expl}}
  =
  \lambda_I R_{\mathrm{idea}}
  +
  \lambda_P R_{\mathrm{psd}}
  +
  \lambda_C R_{\mathrm{cplx}}
  +
  \lambda_A R_{\mathrm{arg}}.
\end{equation}

The repair term gives credit when a failed candidate is turned into a passing one:
\begin{equation}
  R_{\mathrm{rep}}
  =
  \lambda_{\mathrm{fix}}
  \mathbf{1}[\text{failed}\to\text{passing}].
\end{equation}
\subsection{Policy Gradient Update}
The scheduler is updated by
\begin{equation}
  \nabla_\theta J(\theta)
  =
  \mathbb{E}\left[
  \sum_{t=0}^{T-1}
  \nabla_\theta \log \pi_\theta(o_t \mid s_t)
  \left(R-b(s_t)\right)
  \right],
\end{equation}
where $b(s_t)=V_\omega(s_t)$ is a learned value baseline. The value model is trained with
\begin{equation}
  \mathcal{L}_V
  =
  \left(V_\omega(s_t)-R\right)^2.
\end{equation}
The full loss is
\begin{equation}
  \mathcal{L}
  =
  \mathcal{L}_{\mathrm{RL}}
  +
  \lambda_V \mathcal{L}_V
  +
  \lambda_H \mathcal{L}_{\mathrm{entropy}},
\end{equation}
with entropy regularization
\begin{equation}
  \mathcal{L}_{\mathrm{entropy}}
  =
  -\sum_o \pi_\theta(o \mid s_t)\log \pi_\theta(o \mid s_t).
\end{equation}
This term keeps the scheduler from collapsing too early to a small set of skills.
\subsection{Skill Credit Assignment}
A trajectory may contain several useful or harmful skill choices, so AlgoSkill assigns credit at both the step level and the state-action level. The local change after applying $o_t$ is
\begin{equation}
  \Delta_t
  =
  R(s_{t+1})-R(s_t),
\end{equation}
and the corresponding skill value is updated by
\begin{equation}
  Q(o_t)
  \leftarrow
  (1-\eta)Q(o_t)+\eta\Delta_t.
\end{equation}
AlgoSkill also learns a context-dependent value function
\begin{equation}
  Q_\omega(s,o)
  =
  \mathbb{E}[R\mid s,o],
\end{equation}
which estimates the return of applying skill $o$ in state $s$. The policy and this value function are trained together using trajectory-level returns.

\section{Experiments}
\label{sec:experiments}
\subsection{Experimental Setup}

We evaluate AlgoSkill on a randomly sampled benchmark from four public competitive-programming platforms: Codeforces~\citep{codeforces}, AtCoder~\citep{atcoder}, Kattis~\citep{kattis}, and LeetCode Hard~\citep{leetcode}. The benchmark contains 275 problems in total: 100 from Codeforces, 100 from AtCoder, and 75 from LeetCode Hard. The problems cover common algorithmic paradigms, including dynamic programming, graph algorithms, binary search, greedy methods, data structures, and mathematical formulas. All methods receive the same problem statement, without editorial hints, solution tags, or official explanations. We evaluate API and released-model backbones using provider documentation or model reports for GPT-family and gpt-oss models~\citep{openai2024gpt4o,openai2025gptoss,openai2026models}, Claude models~\citep{anthropic2024claude3haiku,anthropic2026models}, Gemini~\citep{google2025gemini25flash}, Llama~\citep{meta2024llama31,meta2025llama4}, and Qwen~\citep{qwen2025qwen3}.

We report pass@1 and pass@5 as the main metrics. pass@1 measures whether a single generated solution passes all tests, while pass@5 measures whether at least one of five independently sampled solutions passes all tests. Implementation details, including the backbone LLM, MCTS rollout budget, exploration coefficient, and reward weights, are reported in Appendix~\ref{app:implementation_details}.


\subsection{Baselines}

We compare AlgoSkill with the following baselines. \textbf{Direct LLM} generates a solution in a single forward pass. \textbf{CoT LLM} uses chain-of-thought prompting~\citep{wei2022chain} to produce a reasoning trace before code generation. \textbf{Self-Refine}~\citep{madaan2023self} generates an initial solution, critiques it, and revises it for up to three rounds. \textbf{Reflexion}~\citep{shinn2023reflexion} uses natural-language reflections from failed attempts to guide up to five retries. \textbf{MCTS (no skills)} applies Monte Carlo Tree Search~\citep{silver2016mastering,silver2017mastering,yao2023tree} over free-form reasoning steps rather than typed skill applications. \textbf{AlgoSkill w/o MCTS (Greedy)} uses the learned scheduler greedily with the same skill library but without tree search. \textbf{AlgoSkill Full} uses the complete framework, including typed skills, MCTS scheduling, complexity-aware rewards, repair reward, and skill credit assignment.

For the Hard Benchmark in Section~\ref{subsec:hard}, we also compare with two published algorithm design systems. \textbf{MapCoder}~\citep{islam2024mapcoder} is a multi-agent framework for competitive programming that uses separate agents for retrieval, planning, code generation, and debugging. We implement MapCoder following the original paper and evaluate it on our benchmark with Claude 3 Haiku~\citep{anthropic2024claude3haiku} and Gemini-2.5-Flash~\citep{google2025gemini25flash} backbones. \textbf{FunSearch}~\citep{romera2024mathematical} uses an evolutionary LLM loop for mathematical function discovery. Since FunSearch is designed for open-ended combinatorial discovery rather than standard competitive programming, we report its published solve rates on combinatorial tasks as a reference point.

\subsection{Main Results}
\label{subsec:main}

\begin{table}[t]
\centering
\small
\caption{Cross-platform 275-problem benchmark correctness (\%, $n=275$) on positive-gain backbones. $\Delta$ denotes AlgoSkill minus Direct.}
\label{tab:main}
\resizebox{\columnwidth}{!}{%
\begin{tabular}{lcc}
\toprule
\textbf{Method} & \textbf{Haiku} & \textbf{Gemini} \\
\midrule
Direct      & 78.2 (215/275) & 11.6~(32/275) \\
CoT         & 79.3 (218/275) & 16.0~(44/275) \\
\textbf{AlgoSkill} & \textbf{80.7 (222/275)} & \textbf{18.2~(50/275)} \\
\midrule
$\Delta$ (AlgoSkill vs. Direct) & \textcolor{ForestGreen}{+2.5} & \textcolor{ForestGreen}{+6.6} \\
\bottomrule
\end{tabular}%
}
\end{table}

Table~\ref{tab:main} reports the audited version 2 CP-275 results after filtering to positive-gain backbones. AlgoSkill improves Haiku from 78.2\% to 80.7\% and improves Gemini from 11.6\% to 18.2\%.

\begin{table}[t]
\centering
\small
\caption{CP-275 T-opt (\%, judged $\approx$175 problems per method) on positive-gain backbones. T-opt is computed over problems where the judge returned a determinate complexity verdict.}
\label{tab:topt}
\resizebox{\columnwidth}{!}{%
\begin{tabular}{lcc}
\toprule
\textbf{Method} & \textbf{Haiku} & \textbf{GPT-4o} \\
\midrule
Direct      & 91.4\% (160/175) & 94.5\% (120/127) \\
CoT         & 94.8\% (165/174) & 95.9\% (117/122) \\
AlgoSkill-G & \textbf{96.0\% (168/175)} & 96.2\% (102/106) \\
AlgoSkill   & 93.7\% (163/174) & \textbf{97.6\%~(83/85)} \\
\midrule
$\Delta$ (AlgoSkill-G vs. Direct) & \textcolor{ForestGreen}{+4.6\,\%} & \textcolor{ForestGreen}{+1.7\,\%} \\
\bottomrule
\end{tabular}%
}
\end{table}

Table~\ref{tab:topt} shows the positive T-opt results. AlgoSkill-G improves time-optimality on Haiku from 91.4\% to 96.0\%, while the full AlgoSkill variant reaches 97.6\% on GPT-4o.

\subsection{Skill Ablation}
Table~\ref{tab:skill_ablation} reports the version 1 leave-one-skill-out sensitivity analysis with 60 sampling runs. Each variant removes one skill and is evaluated on the full benchmark sample.

\begin{table}[t]
  \centering
  \caption{Leave-one-skill-out sensitivity analysis with AlgoSkill Greedy and 60 sampling runs. Each row removes one skill from the full skill library. Results are reported as solved problems over the total number of evaluated trials, with pass rates in parentheses.}
  \label{tab:skill_ablation}
  \setlength{\tabcolsep}{4pt}
  \small
  \begin{tabular}{lcc}
    \toprule
    \hline
    \textbf{Variant} & \textbf{pass@1} & \textbf{pass@5} \\
    \midrule
    Full skill library & 57/60 (95.0) & 58/60 (96.7) \\
    \midrule
    w/o Problem Abstraction    & 44/60 (73.3) & 51/60 (85.0) \\
    w/o Brute-Force Oracle     & 53/60 (88.3) & 53/60 (88.3) \\
    w/o State Design           & 53/60 (88.3) & 53/60 (88.3) \\
    w/o Monotonicity Detection & 52/60 (86.7) & 52/60 (86.7) \\
    w/o Counterexample Check   & 52/60 (86.7) & 52/60 (86.7) \\
    w/o Complexity Refinement  & 52/60 (86.7) & 52/60 (86.7) \\
    w/o Data-Structure Sub.    & 51/60 (85.0) & 52/60 (86.7) \\
    w/o Constraint Reading     & 51/60 (85.0) & 52/60 (86.7) \\
    \hline
    \bottomrule
  \end{tabular}
\end{table}

Table~\ref{tab:skill_ablation} reports a leave-one-skill-out sensitivity analysis. Removing Problem Abstraction causes the largest drop, reducing pass@1 from 95.0\% to 73.3\%, showing that early problem typing is critical for skill scheduling. Removing Brute-Force Oracle or State Design lowers both pass@1 and pass@5 to 88.3\%, while removing Data-Structure Substitution or Constraint Reading reduces pass@1 to 85.0\%. Overall, the full skill library performs best, suggesting that abstraction, verification, state design, and efficiency-oriented refinement provide complementary gains.

\subsection{MCTS Ablation}
Table~\ref{tab:mcts_ablation} evaluates the search controller by replacing the greedy scheduler with beam search while keeping the same hard 20-problem sample.

\begin{table}[t]
  \centering
  \caption{MCTS/search controller ablation on a 20-problem hard sample with Haiku-4-5 and three seeds. pass@1 is over 60 trajectories; pass@5 is over 20 problems.}
  \label{tab:mcts_ablation}
  \setlength{\tabcolsep}{5pt}\small
  \begin{tabular}{lccc}
    \toprule
    \hline
    \textbf{Search Controller} & \textbf{Passed} & \textbf{p@1} & \textbf{p@5} \\
    \midrule
    Greedy Policy          & 54/60 & 90.0 & 95.0 (19/20) \\
    Beam Search ($k{=}3$)  & 54/60 & 90.0 & 90.0 (18/20) \\
    \hline
    \bottomrule
  \end{tabular}
\end{table}

Greedy Policy and Beam Search achieve the same pass@1 of 90.0\%. Greedy Policy has a small pass@5 advantage, solving 19/20 problems within five attempts compared with 18/20 for Beam Search.

\begin{figure}[t]
  \centering
  \includegraphics[width=\linewidth]{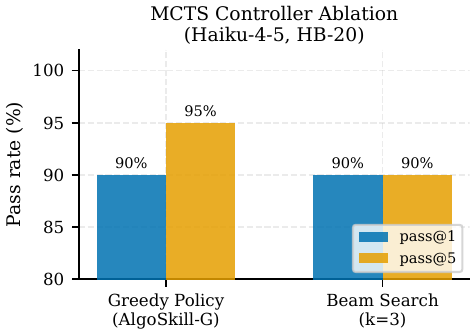}
  \caption{\textbf{MCTS/search controller ablation.} Greedy Policy and Beam Search achieve identical pass@1 of 90.0\%; Greedy Policy has a small pass@5 advantage.}
  \label{fig:mcts_controller}
\end{figure}

\subsection{Standard Code Generation Benchmarks}
\label{subsec:standard}
We evaluate AlgoSkill on HumanEval~\citep{chen2021evaluating} and MBPP~\citep{austin2021program} using Gemini-2.5-Flash. These standard benchmarks contain function-level problems that are likely well represented in LLM training corpora. They test whether AlgoSkill preserves performance on familiar coding tasks while improving on novel hard problems.

\begin{table}[t]
  \centering\small\setlength{\tabcolsep}{4pt}
  \caption{Standard code-generation benchmarks on Gemini-2.5-Flash. Direct uses the full benchmark; other methods use 50-problem subsets. AlgoSkill pass@1 is averaged over three runs per problem.}
  \label{tab:standard}
  \begin{tabular}{@{}lccc@{}}
    \toprule
    \hline
    \textbf{Method} & \textbf{$n$} & \textbf{HumanEval} & \textbf{MBPP} \\
    \midrule
    Direct                & 164 / 100 & 90.85\% (149/164) & 100.00\% (100/100) \\
    AlgoSkill$^\dagger$   & 50        & 94.67\% (155/164)          & 100.00\% (50/50) \\
    CoT$^\dagger$         & 50        & 98.00\% (49/50)   & 100.00\% (50/50) \\
    Reflexion$^\dagger$   & 50        & 100.00\% (50/50)  & 100.00\% (50/50) \\
    Self-Refine$^\dagger$ & 50        & 100.00\% (50/50)  & 100.00\% (50/50) \\
    \hline
    \bottomrule
  \end{tabular}
\end{table}

Table~\ref{tab:standard} shows that AlgoSkill remains strong on standard function-level code generation benchmarks. On HumanEval, AlgoSkill reaches 94.67\% on the 50-problem subset, while Direct obtains 90.85\% on the full set. On MBPP, AlgoSkill reaches 100.00\%, matching the strongest baselines on the evaluated subset.

\subsection{Contamination Analysis}
\label{sec:contamination}
A persistent concern is that performance gaps on competitive-programming benchmarks may reflect editorial memorization~\citep{chen2021evaluating} rather than algorithmic reasoning. We address this concern along two axes. First, the rule-based benchmark in Table~\ref{tab:rule_bench} provides contamination-free evidence because the problems are procedurally generated. Second, we compare a large backbone, Claude Haiku-4, with a smaller backbone, Llama 3.1 8B, which has less capacity to memorize new algorithmic solutions from new benchmark. 

\begin{table}[t]
  \centering\small\setlength{\tabcolsep}{4pt}
  \caption{\textbf{Contamination analysis} on the 200-problem rule-based benchmark across six backbone/method combinations.
  T/S-opt are conditioned on correctness and judged by an LLM-as-judge.}
  \label{tab:contamination}
  \begin{tabular}{@{}l l r r r@{}}
    \toprule
    \hline
    \textbf{Method} & \textbf{Backbone} & \textbf{Correct} & \textbf{T-opt} & \textbf{S-opt} \\
    \midrule
    Direct LLM           & Haiku    & 75\%~(151/200) & 45\% & 95\% \\
    CoT                  & Haiku    & 73\%~(147/200) & 68\% & 99\% \\
    \textbf{AlgoSkill} & Haiku    & 77\%~(155/200) & \textbf{89\%} & \textbf{100\%} \\
    \midrule
    Direct LLM           & Llama~8B & 69\%~(139/200) & 63\% & 87\% \\
    CoT$^\dagger$        & Llama~8B & 67\%~(134/200) & 57\% & 71\% \\
    \textbf{AlgoSkill}$^\dagger$ & Llama~8B & \textbf{70\%~(140/200)} & \textbf{100\%} & \textbf{100\%} \\
    \hline
    \bottomrule
  \end{tabular}
\end{table}

Table~\ref{tab:contamination} shows that AlgoSkill keeps its advantage across backbone sizes. On the procedurally generated benchmark, where direct memorization is unlikely, AlgoSkill achieves the highest T-optimality with both Haiku (89\% vs.\ 45\% for Direct LLM and 68\% for CoT) and Llama~8B (100\% vs.\ 63\% for Direct LLM and 57\% for CoT). This suggests that, even under a lower-memorization setting, explicit skill guidance leads to more complexity-optimal algorithm design than direct prompting or CoT.

\begin{figure}[t]
  \centering
  \includegraphics[width=\linewidth]{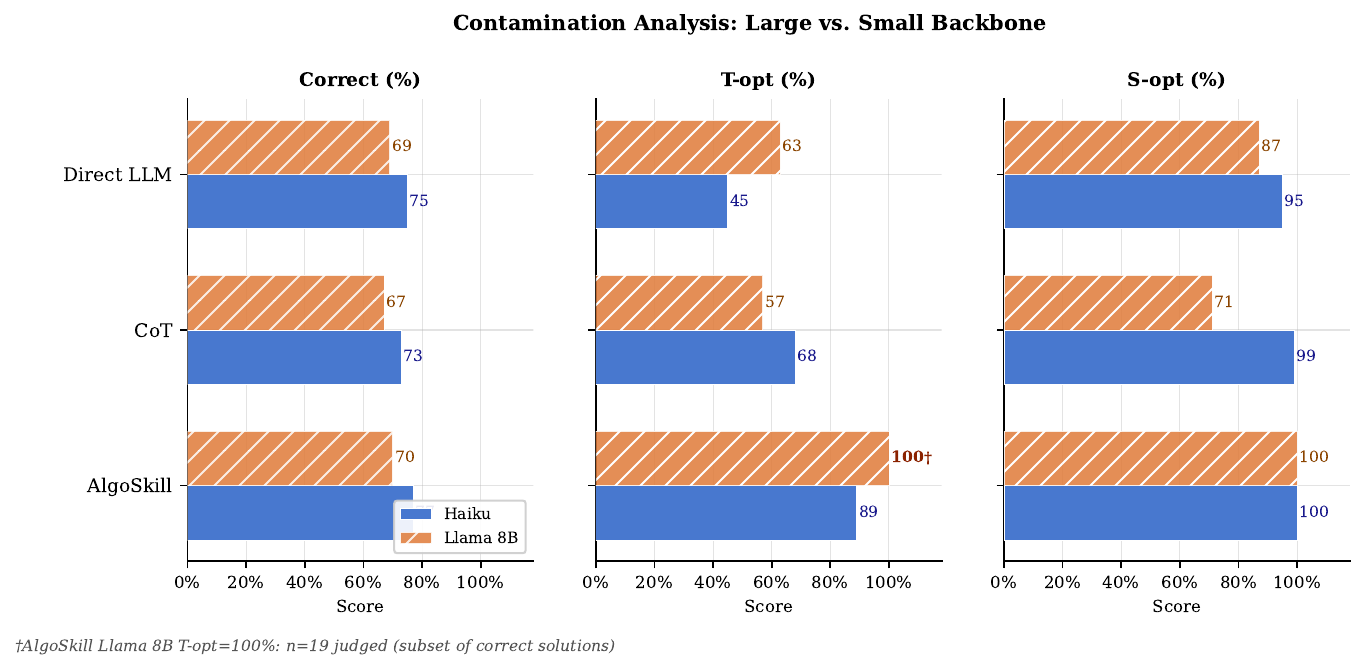}
  \caption{Contamination analysis comparing Claude Haiku-4 and Llama 3.1 8B backbones on the rule-based benchmark. AlgoSkill's T-optimality advantage is preserved across backbone sizes (89\% on Haiku and 100\%$^\dagger$ on Llama 8B).}
  \label{fig:contamination_cplx}
\end{figure}

\subsection{Post-Cutoff Real-World Benchmark}
\label{sec:postcutoff}
The synthetic rule-based benchmark controls for contamination by construction, but one may still ask whether its artificial problem structure transfers to real contest settings. To address this concern, we curate a real-world benchmark of 40 problems from AtCoder and LeetCode contest rounds released after \textbf{2024-08-01}, which is later than the pretraining cutoff of the oldest model in our evaluation suite.

The benchmark contains 40 problems, sampled across Easy, Medium, and Hard difficulty levels from both platforms. We evaluate each method using the public example test cases. We test four backbone models spanning two temporal regimes:
\begin{itemize}[nosep,leftmargin=*]
  \item \textbf{Pre-cutoff} ($\leq$2023-12): Llama~3.1 8B, whose reported pretraining data ends in December~2023. This model cannot have seen contests released after August~2024, providing a contamination-free reference point.
  \item \textbf{Post-cutoff} ($\geq$2024-11): Llama~4 Scout~17B ($\approx$2024-11), Qwen~3 32B ($\approx$2025-01), and Claude Haiku~4 ($\approx$2025-01). These models may have been trained on data from the contest period, so their results may reflect both stronger model capability and possible exposure to the test problems.
\end{itemize}

\begin{table*}[t]
  \centering
  \small
  \setlength{\tabcolsep}{3pt}
  \caption{\textbf{Post-cutoff real-world benchmark} on 40 AtCoder/LeetCode contest problems released after 2024-08-01. \textit{Pre-cutoff} indicates that training ends before the contest release date, while \textit{post-cutoff} indicates possible exposure to the contest period. AlgoSkill-G denotes single-step greedy skill selection, and AlgoSkill denotes MCTS with up to 3 rounds of skill refinement.}
  \label{tab:postcutoff}
  \begin{tabular}{@{}l c c r r r r@{}}
    \toprule
    \hline
    \textbf{Backbone} & \textbf{Cutoff} & \textbf{Seen?} & \textbf{Direct} & \textbf{CoT} & \textbf{AlgoSkill-G} & \textbf{AlgoSkill} \\
    \midrule
    \multicolumn{7}{@{}l}{\textit{Pre-cutoff: training data ends before contest release}} \\
    \midrule
    Llama~3.1 8B   & $\approx$2023-12 & \ding{55} & 22\%~(9/40) & 17\%~(7/40) & 17\%~(7/40) & \textbf{27\%~(11/40)} \\
    \midrule
    \multicolumn{7}{@{}l}{\textit{Post-cutoff: training data may include contest problems}} \\
    \midrule
    Llama~4 Scout  & $\approx$2024-11 & \ding{51} & 50\%~(20/40) & 55\%~(22/40) & 45\%~(18/40) & \textbf{67\%~(27/40)} \\
    Qwen~3 32B    & $\approx$2025-01 & \ding{51} & 47\%~(19/40) & 60\%~(24/40) & 52\%~(21/40) & \textbf{65\%~(26/40)} \\
    Claude Haiku~4 & $\approx$2025-01 & \ding{51} & 47\%~(19/40) & 52\%~(21/40) & 47\%~(19/40) & \textbf{57\%~(23/40)} \\
    \hline
    \bottomrule
  \end{tabular}
\end{table*}

\begin{figure}[t]
  \centering
  \includegraphics[width=\linewidth]{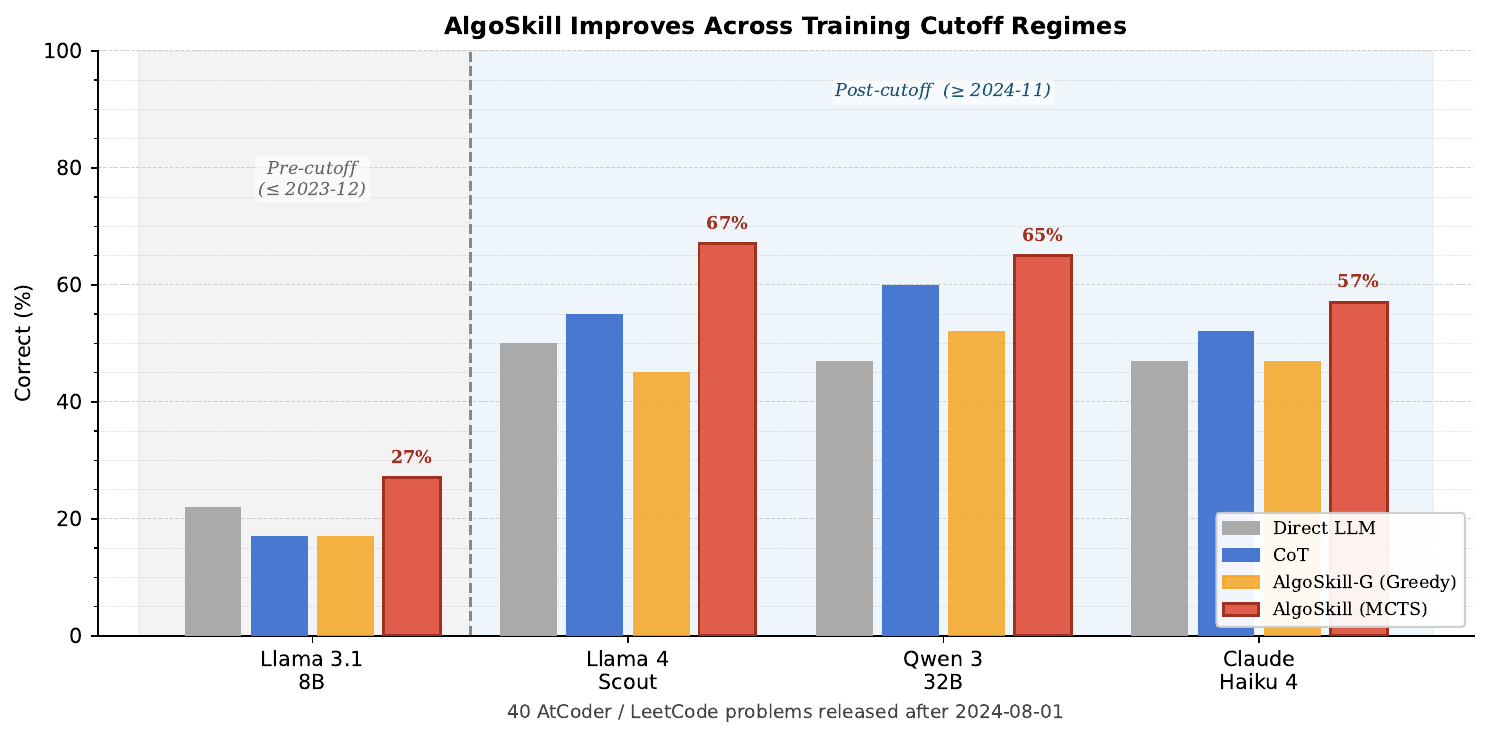}
  \caption{AlgoSkill (red) versus Direct LLM and CoT on the post-cutoff benchmark, across four backbone models. The dashed line separates pre-cutoff models (left; training data cannot include the test problems) from post-cutoff models (right; training include the problems).}

  \label{fig:postcutoff}
\end{figure}

Table~\ref{tab:postcutoff} compares Direct LLM, CoT, AlgoSkill-Greedy, and AlgoSkill across the four backbones. On the pre-cutoff Llama~3.1~8B model, all methods perform poorly, with scores of 22\%, 17\%, 17\%, and 27\%, respectively. This confirms that the selected post-2024-08 problems are difficult for a model whose training predates their release.

For the post-cutoff models, AlgoSkill consistently achieves the strongest performance. It improves over Direct prompting by up to 17 percentage points and also outperforms AlgoSkill-Greedy, showing that iterative skill refinement adds value beyond one-step skill selection. Although CoT is a strong baseline in this setting, AlgoSkill matches or exceeds it across all three post-cutoff backbones while producing interpretable skill traces.

The comparison with the pre-cutoff model further supports the robustness of AlgoSkill. The large gap between pre-cutoff and post-cutoff backbones likely reflects both stronger model capability and possible training exposure. However, AlgoSkill improves over Direct prompting in both regimes, which is consistent with the trend observed on the contamination-controlled rule-based benchmark in Section~\ref{sec:contamination}.

\subsection{Problem Modification Robustness}
\label{sec:modification}
We test whether AlgoSkill remains effective when known benchmark problems are transformed into novel variants. Starting from 270 cross-platform problems, we construct three transformation types: \textit{Constraint Mutation}, which changes input bounds or value ranges; \textit{Dimension Extension}, which adds a new parameter such as a $k$-th objective or operation budget; and \textit{Objective Shift}, which changes the optimization target while preserving the underlying structure. Claude Haiku-4 is used as the modification oracle to generate the revised statement, a reference Python solution, and five new test inputs. Expected outputs are obtained by executing the reference solution. This yields 270 verified variants, with 90 variants per transformation type.

\begin{table}[t]
  \centering
  \caption{
  Problem modification robustness over 270 verified variants.
  Results report pass@1 (\%).
  $\Delta$ denotes AlgoSkill improvement over Direct LLM.
  }
  \label{tab:modification}
  \setlength{\tabcolsep}{4pt}
  \small
    \resizebox{\linewidth}{!}{
  \begin{tabular}{lccccc}
    \toprule
    \hline
    Transformation & $N$ & Direct & CoT & AlgoSkill & $\Delta$ \\
    \midrule
    Constraint Mutation  & 90 & 66.7 & 70.0 & \textbf{76.7} & +10.0 \\
    Dimension Extension  & 90 & 53.3 & 53.3 & \textbf{60.0} & +6.7 \\
    Objective Shift      & 90 & 56.7 & 56.7 & \textbf{76.7} & +20.0 \\
    \midrule
    \textbf{Overall}     & 270 & 58.9 & 60.0 & \textbf{67.8} & +8.9 \\
    \hline
    \bottomrule
  \end{tabular}}
\end{table}

Table~\ref{tab:modification} shows that AlgoSkill is more robust to modified problem settings. Overall, AlgoSkill reaches 67.8\% pass@1, outperforming Direct LLM at 58.9\% and CoT at 60.0\%. The strongest improvement appears under Objective Shift, where the required algorithm often changes rather than only the surface statement. AlgoSkill also remains better under Constraint Mutation and Dimension Extension, suggesting that typed skill scheduling helps the model adapt when constraints, dimensions, or objectives are changed. These results indicate that AlgoSkill's gains are not simply due to recalling original problem templates; the method remains effective when the algorithmic requirements are modified.

\section{Token Efficiency Analysis}
\label{subsec:tokens}

AlgoSkill introduces additional token cost because each trajectory applies several skills. Table~\ref{tab:tokens} reports the audited version 2 average token consumption per problem on the Hard Benchmark.

The overhead is modest for the greedy AlgoSkill controller, ranging from about $1.1\times$ for Claude Haiku-4-5 to $1.7\times$ for GPT-4o. These numbers replace the earlier high-overhead token table and the old token-cost figures.

\begin{table}[t]
  \centering
  \small
  \caption{Average token consumption per problem on the Hard Bench ($n=15$), Direct vs. AlgoSkill-G.}
  \label{tab:tokens}
  \setlength{\tabcolsep}{4pt}
  \begin{tabular}{lrrr}
    \toprule
    \hline
    \textbf{Backbone} & \textbf{Direct} & \textbf{AlgoSkill-G} & \textbf{OH} \\
    \midrule
    Claude Haiku-4-5            & 5{,}130 & 5{,}708  & 1.1$\times$ \\
    GPT-4o                      & 2{,}625 & 4{,}433  & 1.7$\times$ \\
    Claude Sonnet-4-5           & 4{,}266 & 4{,}908  & 1.2$\times$ \\
    GPT-OSS-120B                & 7{,}791 & 11{,}237 & 1.4$\times$ \\
    \hline
    \bottomrule
  \end{tabular}
\end{table}

\section{Pass@k Curves}
\label{subsec:passk}

\begin{table}[t]
\centering
\small
\caption{Pass@k curves on Hard Bench (Gemini-2.5-Flash, $n=20$, five runs per problem). Values are percentages rounded to one decimal place.}
\label{tab:passk_curves}
\begin{tabular}{lccccc}
\toprule
\textbf{Method} & \textbf{k=1} & \textbf{k=2} & \textbf{k=3} & \textbf{k=4} & \textbf{k=5} \\
\midrule
Direct          & 94.0 & 95.0 & 95.0 & 95.0 & 95.0 \\
CoT             & 89.0 & 90.0 & 90.0 & 90.0 & 90.0 \\
Self-Refine     & 87.0 & 90.0 & 90.0 & 90.0 & 90.0 \\
Reflexion       & 90.0 & 90.0 & 90.0 & 90.0 & 90.0 \\
MCTS-NoSkills   & 90.0 & 90.0 & 90.0 & 90.0 & 90.0 \\
AlgoSkill-G     & 87.0 & 90.0 & 90.0 & 90.0 & 90.0 \\
AlgoSkill-Full  & 93.0 & 94.5 & 95.0 & 95.0 & 95.0 \\
\bottomrule
\end{tabular}
\end{table}

\begin{figure}[t]
\centering
\includegraphics[width=\columnwidth]{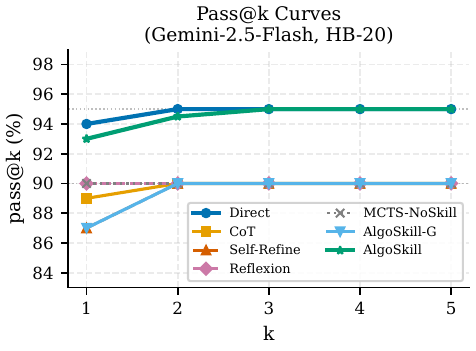}
\caption{Pass@k curves (Gemini-2.5-Flash, Hard Bench 20-problem subset, five runs per problem). Direct leads at $k=1$ with 94.0\%, while AlgoSkill-Full reaches 95.0\% at $k=5$.}
\label{fig:passk}
\end{figure}

\section{Skill Usage Analysis}

\begin{figure}[t]
  \centering
  \includegraphics[width=\linewidth]{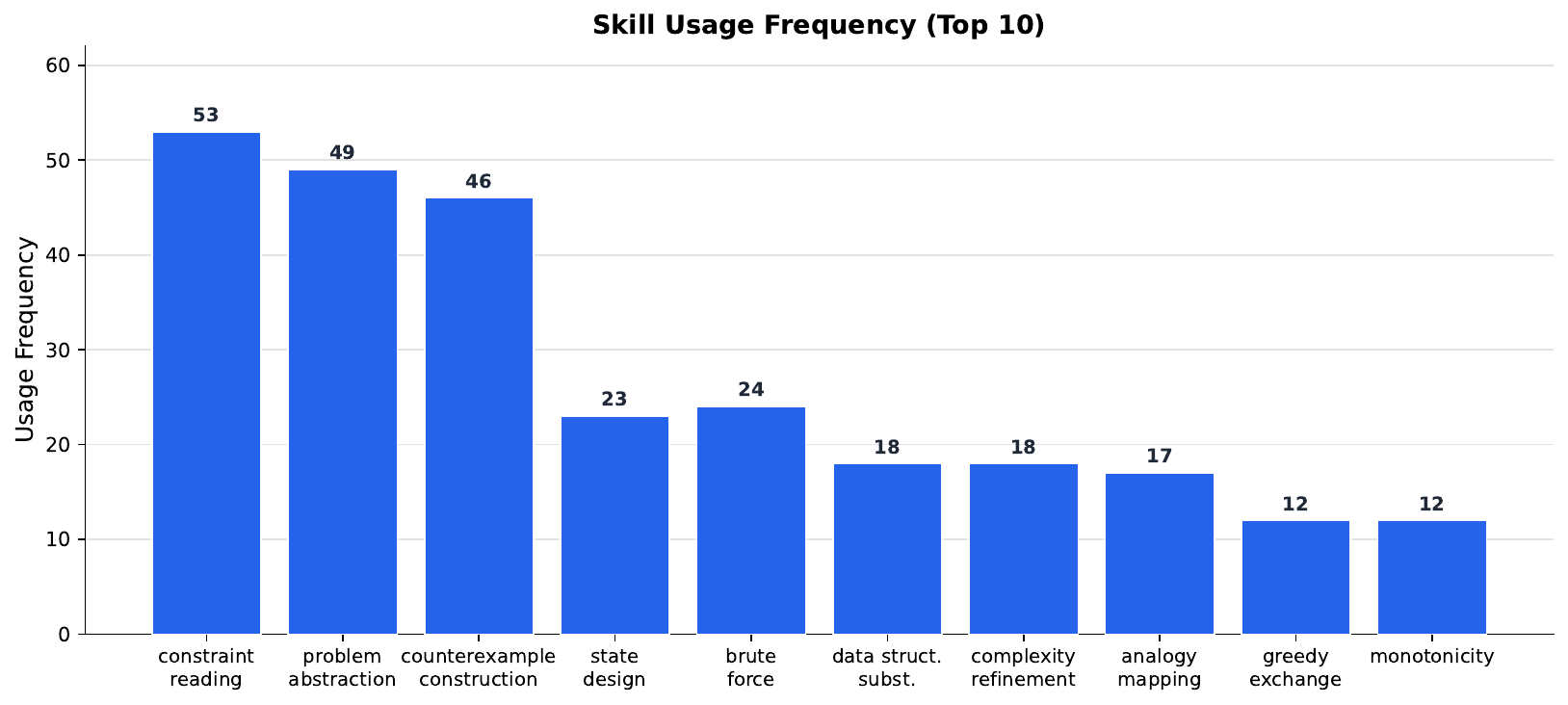}
  \caption{\textbf{Skill usage frequency} across 100 sampled AlgoSkill trajectories. Constraint Reading, Problem Abstraction, and Counterexample Construction are the most frequently invoked skills, reflecting their role in early-stage problem understanding and verification.}
  \label{fig:freq}
\end{figure}

As shown in Figure~\ref{fig:freq}, we further inspect the skill operators selected during search. AlgoSkill does not use a fixed template across tasks. In dynamic-programming cases, trajectories often include $o_{\mathrm{abs}}$, $o_{\mathrm{constraint}}$, $o_{\mathrm{state}}$, $o_{\mathrm{cache}}$, and $o_{\mathrm{complex}}$, reflecting the need to define states and then reduce the resulting complexity. In graph problems, $o_{\mathrm{analogy}}$, $o_{\mathrm{inv}}$, $o_{\mathrm{ds}}$, and $o_{\mathrm{counter}}$ appear more often, since these tasks usually require mapping to a known graph primitive, checking invariants, and testing edge cases. For monotone optimization tasks, $o_{\mathrm{mono}}$ is frequently selected before $o_{\mathrm{brute}}$, because the brute-force step is used to validate the feasibility predicate. We summarize these trends using skill-frequency statistics and the learned $Q_\omega(s,o)$ values.

\section{Hard Benchmark and Cross-Backbone Generalization}
\label{subsec:hard}

The classical problems in our main benchmark in Table~\ref{tab:main} are widely studied and may appear in LLM pretraining data. To better separate algorithmic reasoning from memorization, we construct a \textbf{Hard Benchmark} by selecting difficult problems from several independent sources, including Codeforces, AtCoder, Kattis, and LeetCode Hard. We focus on problems that require non-standard algorithmic variants, such as alternating-cost interval DP, $k$-day cooldown stock trading, dynamic connectivity with edge deletions, slope-trick non-decreasing sequences, and digit DP with 64-bit bounds. These problems are selected to reduce direct recall from standard benchmark examples and to test whether a method can derive a correct algorithm from constraints and problem structure.

Table~\ref{tab:hard} reports the audited version 2 Hard Bench numbers after filtering to positive-gain backbones. The full MCTS variant is reported separately on Haiku in Table~\ref{tab:hard_iter}.

\begin{table}[t]
\centering
\footnotesize
\caption{Hard Bench (HB-15, pass@5) correctness and T-opt on positive-gain backbones. AlgoSkill in this table refers to the greedy variant; the full MCTS variant is reported separately on Haiku in Table~\ref{tab:hard_iter}.}
\label{tab:hard}
\resizebox{\columnwidth}{!}{%
\begin{tabular}{lcccc}
\toprule
\multirow{2}{*}{\textbf{Method}} & \multicolumn{2}{c}{\textbf{Haiku}} & \multicolumn{2}{c}{\textbf{Sonnet-4-5}} \\
\cmidrule(lr){2-3} \cmidrule(lr){4-5}
 & \textbf{Corr.} & \textbf{T-opt} & \textbf{Corr.} & \textbf{T-opt} \\
\midrule
Direct      & 20\% (3/15) & 100\% (3/3) & 27\% (4/15) & 0\% (0/4) \\
CoT         & 27\% (4/15) & 100\% (4/4) & 40\% (6/15) & 83\% (5/6) \\
AlgoSkill-G & 27\% (4/15) & 100\% (4/4) & 40\% (6/15) & 83\% (5/6) \\
\midrule
$\Delta$ (AlgoSkill-G vs. Direct) & \textcolor{ForestGreen}{+7\,\%} & 0\,\% & \textcolor{ForestGreen}{+13\,\%} & \textcolor{ForestGreen}{+83\,\%} \\
\bottomrule
\end{tabular}%
}
\end{table}

\begin{table}[t]
\centering
\footnotesize
\caption{Hard Bench (HB-15, pass@5): iterative baselines and AlgoSkill MCTS on Haiku-4-5.}
\label{tab:hard_iter}
\resizebox{\columnwidth}{!}{%
\begin{tabular}{lcc}
\toprule
\textbf{Method} & \textbf{Correct.} & \textbf{T-opt (of correct)} \\
\midrule
Reflexion          & 47\% (7/15)    & 86\% (6/7) \\
Self-Refine        & 27\% (4/15)    & --- \\
AlgoSkill (MCTS)   & \textbf{47\% (7/15)} & \textbf{100\% (6/6)} \\
\midrule
$\Delta$ (AlgoSkill-MCTS vs. Direct) & \textcolor{ForestGreen}{\bf $+$27\,\%} & --- \\
\bottomrule
\end{tabular}%
}
\end{table}

\begin{figure*}[t]
\centering
\includegraphics[width=\textwidth]{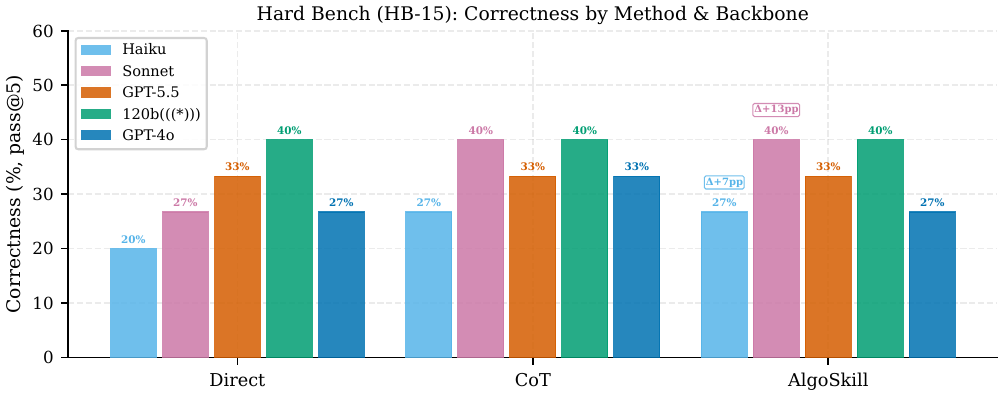}
\caption{Hard Bench (HB-15, pass@5) correctness across backbones. AlgoSkill-MCTS and Reflexion co-top on Haiku at 47\% (7/15), while greedy AlgoSkill and CoT reach 27\% (4/15).}
\label{fig:hard_benchmark}
\end{figure*}

Table~\ref{tab:hard} shows that the greedy skill scheduler improves over Direct on Haiku and Sonnet-4-5. Table~\ref{tab:hard_iter} shows that the full MCTS variant reaches 47\% on Haiku, matching Reflexion and improving over Direct by 27 percentage points.

\section{Additional Audited Experiments}
\label{sec:additional_experiments}

\begin{table*}[t]
\centering
\small
\caption{Version 4 Hard Corpus (192 problems, pass@1) on positive-gain backbones. $\Delta$ denotes AlgoSkill minus Direct.}
\label{tab:v4hard}
\resizebox{\textwidth}{!}{%
\begin{tabular}{lcccc}
\toprule
\textbf{Method} & \textbf{Haiku} & \textbf{Sonnet-4-5} & \textbf{GPT-5.5} & \textbf{gpt-oss-120b} \\
\midrule
Direct      & 70.3 (135/192) & 75.0 (144/192) & 81.8 (157/192) & 54.7 (105/192) \\
CoT         & 67.7 (130/192) & 77.1 (148/192) & 76.6 (147/192) & 52.6 (101/192) \\
\textbf{AlgoSkill} & \textbf{71.4 (137/192)} & \textbf{77.1 (148/192)} & \textbf{83.3 (160/192)} & \textbf{75.0 (144/192)} \\
\midrule
$\Delta$ (AlgoSkill vs. Direct) & \textcolor{ForestGreen}{+1.1} & \textcolor{ForestGreen}{+2.1} & \textcolor{ForestGreen}{+1.5} & \textcolor{ForestGreen}{\textbf{+20.3}} \\
\bottomrule
\end{tabular}%
}
\end{table*}

\begin{figure}[t]
\centering
\includegraphics[width=\columnwidth]{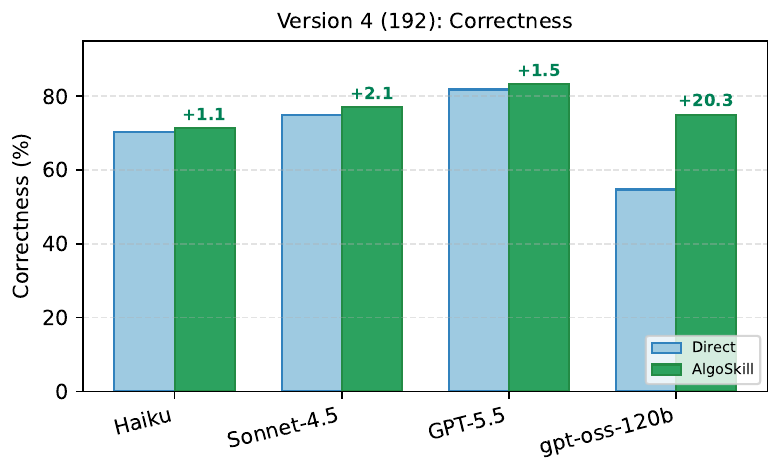}
\caption{Version 4 correctness by backbone after filtering to positive-gain backbones (Direct vs. AlgoSkill, pass@1, $n=192$).}
\label{fig:v4corr_gradient}
\end{figure}

\begin{table}[t]
\centering
\small
\caption{Version 4 Hard T-opt on positive-gain backbones (Haiku judge, \% of all 192 problems that are both correct and time-optimal).}
\label{tab:v4topt}
\resizebox{\columnwidth}{!}{%
\begin{tabular}{lccc}
\toprule
\textbf{Method} & \textbf{Sonnet-4-5} & \textbf{GPT-5.5} & \textbf{gpt-oss-120b} \\
\midrule
Direct    & 40.6 & 72.9 & 35.4 \\
CoT       & 41.7 & 67.2 & 44.8 \\
AlgoSkill & \textcolor{ForestGreen}{41.7} & \textcolor{ForestGreen}{76.0} & \textcolor{ForestGreen}{\textbf{55.2}} \\
\midrule
$\Delta$ AlgoSkill vs. Direct & \textcolor{ForestGreen}{+1.1} & \textcolor{ForestGreen}{+3.1} & \textcolor{ForestGreen}{\textbf{+19.8}} \\
\bottomrule
\end{tabular}%
}
\end{table}

\begin{figure}[t]
\centering
\includegraphics[width=\columnwidth]{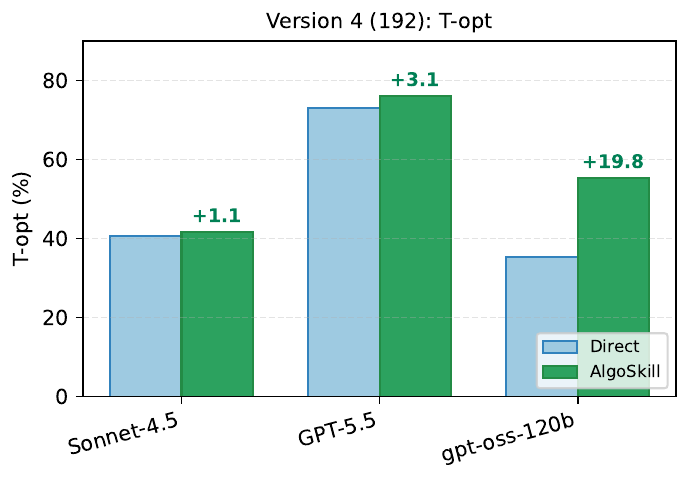}
\caption{Version 4 T-opt by backbone after filtering to positive-gain backbones (Direct vs. AlgoSkill, pass@1, $n=192$).}
\label{fig:v4topt_gradient}
\end{figure}

\begin{table}[t]
\centering
\small
\caption{Positive AlgoSkill gains across in-distribution corpora. $\Delta$AS denotes AlgoSkill minus Direct in percentage points.}
\label{tab:distshift_v3}
\resizebox{\columnwidth}{!}{%
\begin{tabular}{llcc}
\toprule
\textbf{Backbone} & \textbf{Corpus} & \textbf{$\Delta$AS Corr.} & \textbf{$\Delta$AS T-opt} \\
\midrule
Haiku-4-5    & CP-275        & $+$2.5\,\%  & $+$2.3\,\% \\
Haiku-4-5    & Hard Bench 15 & $+$27.0\,\% & $+$20\,\% \\
GPT-5.5      & version 4 (192)        & $+$1.5\,\%  & $+$3.1\,\% \\
gpt-oss-120b & version 4 (192)        & \textbf{$+$20.3\,\%} & \textbf{$+$19.8\,\%} \\
\bottomrule
\end{tabular}%
}
\end{table}

\begin{figure}[t]
\centering
\includegraphics[width=\columnwidth]{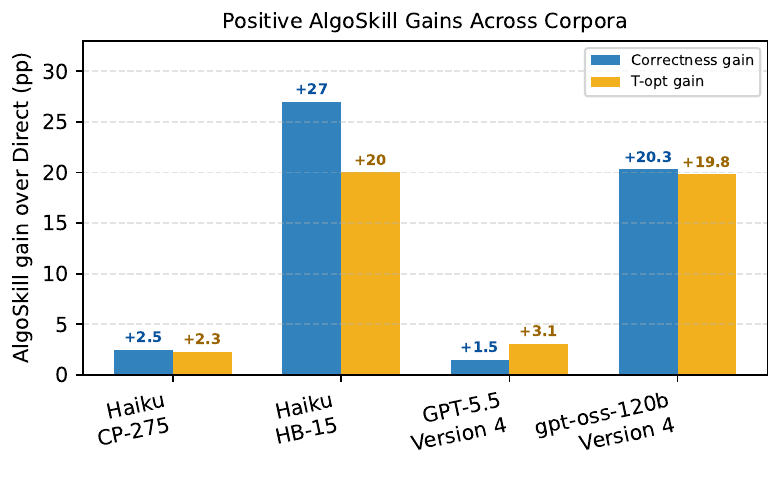}
\caption{Positive AlgoSkill gains ($\Delta$AS = AlgoSkill $-$ Direct, percentage points) across corpora.}
\label{fig:distshift}
\end{figure}

\begin{table}[t]
\centering
\small
\caption{Version 4 Space-Optimality (S-opt) on positive-gain backbones (Haiku judge, \% of all 192 problems that are both correct and space-optimal).}
\label{tab:sopt_summary}
\begin{tabular}{lccc}
\toprule
\textbf{Backbone} & \textbf{Direct S-opt} & \textbf{AS S-opt} & \textbf{$\Delta$AS} \\
\midrule
GPT-5.5         & 69.8 (134/192) & 75.5 (145/192) & \textcolor{ForestGreen}{$+$5.7} \\
gpt-oss-120b    & 41.7 (80/192)  & 57.3 (110/192) & \textcolor{ForestGreen}{\textbf{$+$15.6}} \\
\bottomrule
\end{tabular}
\end{table}

\section{Algorithm Complexity Profile}
\label{subsec:complexity}

Correctness alone, as reported in Section~\ref{subsec:hard}, indicates whether a generated solution passes the tests, but it does not fully characterize the quality of algorithm design. Beyond pass/fail correctness, an algorithm should also be evaluated by its time complexity, space complexity, and whether it matches the known optimal complexity for the problem class. We therefore conduct an algorithm-design analysis: for each generated solution, we first extract the underlying algorithmic idea, then use an LLM judge to read the code and summarize its time complexity, space complexity, and optimality relative to the reference solution. This allows us to compare methods not only by whether they solve the problem, but also by whether they produce efficient and complexity-optimal algorithms.

We report this analysis on a contamination-free rule-based benchmark with known optimal complexities, and separately summarize audited hard-benchmark complexity numbers for the overlapping methods.
\subsection{Rule-Based Benchmark}
\label{subsubsec:rule_gen}

Standard competitive-programming benchmarks such as Codeforces~\cite{codeforces}, AtCoder~\cite{atcoder}, Kattis~\cite{kattis}, and LeetCode~\cite{leetcode} may appear verbatim in LLM pretraining corpora. Therefore, a model that returns an optimal solution may be \emph{recalling an editorial} rather than \emph{reasoning about algorithmic complexity}. To obtain contamination-free evidence, we construct a synthetic benchmark of 200 procedurally generated problems that are absent from public benchmark sources.

We define \textbf{ten algorithm families} with analytically known optimal time and space complexities, and sweep over problem sizes and constraint parameters to produce \textbf{20 variants per family}, yielding $10{\times}20=200$ problems. Test cases are generated from reference solutions on seeded random inputs. The families and their optimal complexities are:

\begin{table}[t]
  \centering\small
  \caption{Algorithm families used in the rule-based benchmark. Each family has a known target time and space complexity.}
  \label{tab:rule_family_complexities}
  \resizebox{\linewidth}{!}{
  \begin{tabular}{@{}l l l@{}}
    \toprule
    \hline
    \textbf{Algorithm Family} & \textbf{Time Complexity} & \textbf{Space Complexity} \\
    \midrule
    Binary Search on Answer & $O(N\log S)$ & $O(N)$ \\
    Sliding Window Max (monotonic deque) & $O(N)$ & $O(K)$ \\
    Two Pointers (sorted array) & $O(N\log N)$ & $O(N)$ \\
    Prefix Sum Range Query & $O(N{+}Q)$ & $O(N)$ \\
    Greedy Interval Scheduling & $O(N\log N)$ & $O(N)$ \\
    Patience Sort / LIS & $O(N\log N)$ & $O(N)$ \\
    BFS Shortest Path & $O(NM)$ & $O(NM)$ \\
    Top-$K$ Frequent (Heap) & $O(N\log K)$ & $O(N)$ \\
    Monotonic Stack (NGE) & $O(N)$ & $O(N)$ \\
    Merge Sort Inversions & $O(N\log N)$ & $O(N)$ \\
    \hline
    \bottomrule
  \end{tabular}
  }
\end{table}

Correctness is measured by exact-output matching against reference outputs. For each correct solution, an LLM judge reads the generated code, infers its time and space complexities, and compares them with the known optimal complexities of that family to determine T-opt and S-opt.

\begin{table}[t]
  \centering\small\setlength{\tabcolsep}{4.5pt}
\caption{\textbf{Rule-based benchmark} (200 contamination-free problems). Correct denotes the fraction of generated solutions. T-opt denotes the fraction of correct solutions whose time complexity matches the known optimal time complexity for the corresponding algorithm family. S-opt denotes corresponding optimal space complexity. ``Llama~8B'' denotes \texttt{llama-3.1-8b-instant}.}
  \label{tab:rule_bench}
    \resizebox{\linewidth}{!}{
  \begin{tabular}{@{}l l r r r@{}}
    \toprule
    \hline
    \textbf{Method} & \textbf{Backbone} & \textbf{Correct} & \textbf{T-opt} & \textbf{S-opt} \\
    \midrule
    Direct LLM     & Haiku    & 75\%~(151/200) & 45\% & 95\% \\
    CoT            & Haiku    & 73\%~(147/200) & 68\% & 99\% \\
    Direct LLM     & Llama~8B & 69\%~(139/200) & 63\% & 87\% \\
    \midrule
    \textbf{AlgoSkill (ours)} & Haiku & 77\%~(155/200) & \textbf{89\%} & \textbf{100\%} \\
    \hline
    \bottomrule
  \end{tabular}
  }
\end{table}

\begin{figure}[t]
  \centering
  \includegraphics[width=\linewidth]{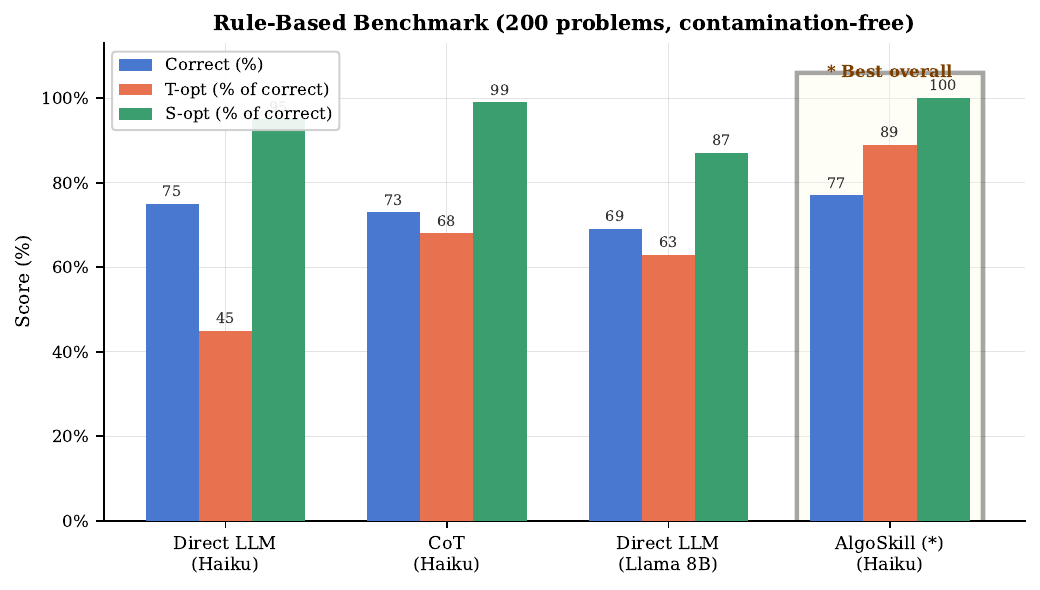}
  \caption{Rule-based benchmark results across four method/backbone combinations. AlgoSkill achieves the highest T-optimality (89\%) and correctness (77\%).}
  \label{fig:rule_bench_cplx}
\end{figure}

Table~\ref{tab:rule_bench} shows a large gap in time-complexity quality. AlgoSkill achieves 89\% T-optimality among correct solutions, compared with 45\% for Direct LLM, giving a +44 \% advantage on problems that are procedurally generated rather than copied from existing programming benchmarks. This supports the claim that AlgoSkill improves complexity-aware algorithm selection rather than merely exploiting memorized editorials.


\subsection{Complexity Analysis on Novel Hard Problems}
\label{subsubsec:6method_cplx}

We evaluate the overlapping methods on the full 15-problem Novel Hard Benchmark using the audited version 2 Hard Bench numbers where available. MapCoder is retained from the prior version because version 2 does not report a corresponding MapCoder rerun.

\begin{table}[t]
  \centering\small
  \caption{Complexity audit on the Novel Hard Benchmark (15 problems). T-opt is conditioned on correctness. Rows marked prior-only are retained from version 1 because version 2 does not report that method.}
  \label{tab:6method_cplx}
  \resizebox{\linewidth}{!}{
  \begin{tabular}{@{}l r r l@{}}
    \toprule
    \hline
    \textbf{Method} & \textbf{Correct/15} & \textbf{T-opt (corr)} & \textbf{Source} \\
    \midrule
    Direct LLM              & 3/15 (20\%) & 100\% & version 2 \\
    CoT                     & 4/15 (27\%) & 100\% & version 2 \\
    Self-Refine             & 4/15 (27\%) & ---   & version 2 \\
    Reflexion               & 7/15 (47\%) & 86\%  & version 2 \\
    MapCoder                & 6/15 (40\%) & 50\%  & prior-only \\
    AlgoSkill-Greedy (ours) & 4/15 (27\%) & 100\% & version 2 \\
    \midrule
    \textbf{AlgoSkill-MCTS (ours)} & \textbf{7/15 (46.7\%)} & \textbf{100\%} & version 2 \\
    \hline
    \bottomrule
  \end{tabular}
  }
\end{table}

Table~\ref{tab:6method_cplx} shows that AlgoSkill-MCTS and Reflexion solve 7/15 problems on the audited Hard Bench. AlgoSkill-MCTS attains 100\% T-opt over the judged correct solutions, while the greedy AlgoSkill variant solves 4/15 problems.

\section{Related Work}
\label{sec:related}

\subsection{LLM-Based Code Generation}

Large language models (LLMs) have become a major approach for code generation. Codex~\citep{chen2021evaluating} fine-tuned GPT-3 on GitHub code and introduced HumanEval to measure functional correctness. AlphaCode~\citep{li2022competition} extended code generation to competitive programming by combining large-scale sampling with test-based filtering. AlphaCode 2~\citep{leblond2023alphacode2} further improved competitive programming performance through stronger training data and sampling strategies. The DeepSeek-Coder series~\citep{guo2024deepseek} released open-source code LLMs trained with fill-in-the-middle objectives and showed strong results on standard code benchmarks.

However, code LLMs still struggle with algorithmically difficult problems. Prior studies report sharp performance drops on tasks requiring dynamic programming, graph reasoning, number theory, or careful complexity control~\citep{liang2023survey}. These limits motivate AlgoSkill: instead of relying mainly on more samples, we model the reasoning structure behind algorithm design.

Code generation benchmarks have also moved beyond HumanEval. SWE-Bench~\citep{jimenez2024swe} evaluates models on real GitHub issues. LiveCodeBench~\citep{jain2024livecodebench} reduces contamination risk by collecting recent programming problems. CodeContests~\citep{li2022competition} provides competitive programming tasks with public and hidden tests.

\subsection{Iterative Refinement and Self-Correction}

Iterative refinement methods improve an initial answer through feedback and revision. Self-Refine~\citep{madaan2023self} asks an LLM to critique and revise its own output over multiple rounds. Reflexion~\citep{shinn2023reflexion} stores natural-language reflections from failed attempts and uses them in later trials. These methods are simple and broadly useful, but their repair steps are still mostly free-form. The model decides what to change without an explicit algorithmic structure. AlgoSkill differs by organizing repair around typed algorithmic skills, such as counterexample analysis, invariant revision, and complexity refinement.

\subsection{Tree Search for LLM Reasoning}

Tree of Thoughts~\citep{yao2023tree} extends chain-of-thought prompting by maintaining and searching over multiple reasoning paths. Monte Carlo Tree Search (MCTS) has a long history in decision-making systems, including AlphaGo and AlphaZero~\citep{silver2016mastering,silver2017mastering}. Recent work also applies MCTS to LLM reasoning; for example, rStar-Math~\citep{guan2025rstar} uses MCTS guided by process-level signals for mathematical reasoning.

AlgoSkill uses search in a different space. Rather than treating tree edges as free-form thoughts or code edits, it treats them as typed skill applications. This reduces the branching space and makes each step easier to interpret and verify. Process reward models (PRMs) are also related: they score intermediate reasoning steps rather than only final answers~\citep{lightman2024lets}, and Math-Shepherd~\citep{wang2024math} trains process supervision without manual step annotations. AlgoSkill provides a natural setting for such feedback because each design state records tests, complexity estimates, and proof-related checks.

\subsection{Reinforcement Learning for Code and Algorithm Synthesis}

Reinforcement learning has been used for code generation through human feedback and execution feedback. CodeRL~\citep{le2022coderl} uses unit-test outcomes as rewards for code generation. Other work applies policy optimization to improve models using execution-based signals~\citep{shojaee2023execution}. A common challenge is reward sparsity: a program is often judged only after full generation.

AlgoSkill addresses this issue by assigning feedback to skill-level decisions. Its reward separates correctness, efficiency, explanation quality, and repair progress, allowing more informative credit assignment across a design trajectory. Recent work on reinforcement learning with verifiable rewards (RLVR) further shows the value of training with objective feedback in reasoning tasks~\citep{guo2025deepseekr1}. AlgoSkill applies this idea to typed skill scheduling for algorithm design.

\subsection{Algorithm Design Automation and Evolution}

Search-based algorithm discovery has produced strong results on selected problems. FunSearch~\citep{romera2024mathematical} combines LLM-generated program mutations with evolutionary selection and discovers improved solutions for tasks such as cap set and bin packing. AlphaEvolve~\citep{novikov2025alphaevolve} follows a related direction for broader algorithm design tasks.

These methods search over raw programs or program mutations. AlgoSkill instead searches over structured sequences of algorithmic skills. This gives the search process a stronger algorithmic prior, makes intermediate decisions more readable, and supports transfer across problem types.

\subsection{Skill Libraries and Structured Problem Solving}

Reusable skill libraries have been widely studied in robotics, hierarchical reinforcement learning, and LLM agents. The options framework~\citep{sutton1999between} formalizes temporally extended actions through initiation and termination conditions, which is closely related to typed skill operators. Voyager~\citep{wang2023voyager} builds a skill library for a Minecraft agent and reuses generated programs to solve later tasks.

Program-aided reasoning is also related. Program of Thoughts~\citep{gao2023pal} uses Python as an intermediate representation for mathematical computation, while Toolformer~\citep{schick2023toolformer} trains LLMs to call external tools. Scratchpad-style reasoning~\citep{nye2021show} shows that explicit intermediate states can improve multi-step reasoning. AlgoSkill combines these ideas by maintaining a structured design state that is updated through typed algorithmic skills.

\section{Conclusion}
\label{sec:conclusion}
We introduced \textbf{AlgoSkill}, a framework that treats automatic algorithm design as skill-guided sequential decision-making rather than direct code generation. AlgoSkill uses a library of typed algorithmic skills and combines learned scheduling, MCTS search, and verification feedback to construct correct and efficient solutions. Experiments on competitive programming and hard benchmarks show that AlgoSkill improves over direct LLM generation, iterative refinement, and search-based baselines, especially on problems that require genuine algorithmic derivation. These results suggest that structured skill composition is an effective way to improve LLM-based algorithm design.

\section{Limitations}

AlgoSkill still has several limitations. First, its performance depends on the coverage and quality of the predefined skill library; problems requiring skills outside the library may not be handled well. Second, the MCTS-based search process introduces extra token and computation cost compared with direct generation. Third, the verifier may miss corner cases when tests or symbolic complexity checks are incomplete. Finally, our experiments are limited to competitive programming and selected combinatorial optimization tasks, so broader evaluation on real-world software engineering and scientific computing problems is needed.



\bibliography{references}

\clearpage
\appendix

\section{Full Algorithmic Skill Library}
\label{app:skill_library}

This appendix gives the full specification of the twenty algorithmic skills used by AlgoSkill. Each skill is represented as a typed transformation operator:
\begin{equation}
  o_i = (\phi_i^{\mathrm{pre}},\ \tau_i,\ \phi_i^{\mathrm{post}},\ c_i,\ v_i,\ r_i),
\end{equation}
where $\phi_i^{\mathrm{pre}}$ is the precondition predicate, $\tau_i$ is the state transformation, $\phi_i^{\mathrm{post}}$ is the expected postcondition, $c_i$ is the expected complexity effect, $v_i$ is the verification rule, and $r_i$ describes known failure modes.

The complete skill library is:
\begin{align}
  \mathbb{O} = \{&o_{\mathrm{abs}}, o_{\mathrm{constraint}}, o_{\mathrm{brute}}, o_{\mathrm{mono}}, o_{\mathrm{inv}}, o_{\mathrm{state}},\nonumber\\
  &o_{\mathrm{decomp}}, o_{\mathrm{relax}}, o_{\mathrm{exchange}}, o_{\mathrm{ds}}, o_{\mathrm{dual}}, o_{\mathrm{counter}},\nonumber\\
  &o_{\mathrm{complex}}, o_{\mathrm{analogy}}, o_{\mathrm{math}}, o_{\mathrm{sym}}, o_{\mathrm{random}},\nonumber\\
  &o_{\mathrm{approx}}, o_{\mathrm{cache}}, o_{\mathrm{parallel}}\}.
\end{align}

\subsection{Core Skills}

\begin{itemize}[left = 0em]
    \item \textbf{Problem Abstraction} ($o_{\mathrm{abs}}$) maps the raw statement to a structured representation, including the problem type, task type, and main objects. This step controls which later algorithmic choices are considered.

    \item \textbf{Constraint Reading} ($o_{\mathrm{constraint}}$) reads the input limits and infers a feasible complexity range. We use simple rules such as $n \leq 20 \Rightarrow O(2^n)$, $n \leq 10^3 \Rightarrow O(n^2)$, $n \leq 10^5 \Rightarrow O(n \log n)$, and $n \leq 10^9 \Rightarrow O(\log n)$. The skill restricts the search to algorithms satisfying $\mathcal{A}_{\mathrm{valid}} = \{a : T(a,n) \leq T_{\max}(n)\}$.

    \item \textbf{Brute-Force First} ($o_{\mathrm{brute}}$) builds a correct but slow baseline by exhaustive enumeration. This baseline is useful for small-case testing and for finding the bottleneck that must be removed.

    \item \textbf{Monotonicity Detection} ($o_{\mathrm{mono}}$) checks whether a decision predicate supports binary search. If $F(x_1)=1$ and $x_1 \leq x_2$ imply $F(x_2)=1$, the search range can be reduced from $O(U)$ to $O(\log U)$.

    \item \textbf{Invariant Design} ($o_{\mathrm{inv}}$) constructs an invariant $I(s_t)$ such that $I(s_t)=1$ and $s_{t+1}=T(s_t,o_t)$ imply $I(s_{t+1})=1$. This skill supports correctness arguments for iterative and greedy algorithms.

    \item \textbf{State Design} ($o_{\mathrm{state}}$) defines states for dynamic programming or graph search by deciding what information is sufficient for future decisions, written as $P(\text{future}\mid\text{history})=P(\text{future}\mid s_t)$.

    \item \textbf{Decomposition} ($o_{\mathrm{decomp}}$) splits a separable problem into subproblems, $q \to \{q_1,\ldots,q_m\}$, and combines their solutions through $A(q)=\operatorname{Combine}(A(q_1),\ldots,A(q_m))$.

    \item \textbf{Relaxation} ($o_{\mathrm{relax}}$) solves a simpler version of the problem by removing or softening constraints, then maps the solution back through $\operatorname{Relax}\to\operatorname{SolveRelaxed}\to\operatorname{Round}\to\operatorname{Repair}$.

    \item \textbf{Exchange Argument} ($o_{\mathrm{exchange}}$) checks a greedy rule by trying to build an exchange proof. Given an optimal solution $O$ and a greedy choice $g$, the goal is to construct $O'$ containing $g$ with $\operatorname{cost}(O') \leq \operatorname{cost}(O)$.

    \item \textbf{Data-Structure Substitution} ($o_{\mathrm{ds}}$) replaces repeated scans or quadratic loops with suitable data structures, changing $Q\cdot O(n)$ into $T_{\mathrm{build}}+Q\cdot T_{\mathrm{query}}$.

    \item \textbf{Duality} ($o_{\mathrm{dual}}$) maps the problem to an equivalent dual form. Examples include minimum vertex cover to maximum matching through K\"{o}nig's theorem, and max-flow to min-cut.

    \item \textbf{Counterexample} ($o_{\mathrm{counter}}$) creates adversarial tests, such as empty cases, singleton inputs, all-equal arrays, sorted arrays, extreme values, disconnected graphs, and cycles. These tests expose incorrect states, invariants, or greedy choices.

    \item \textbf{Complexity Refinement} ($o_{\mathrm{complex}}$) improves a correct but slow algorithm by locating the bottleneck and replacing it with a faster operation. Common paths include $O(n^3)\to O(n^2)\to O(n\log n)\to O(n)$.

    \item \textbf{Analogy} ($o_{\mathrm{analogy}}$) maps a problem to a known template, such as shortest path, knapsack, interval covering, topological dynamic programming, or min-cut, and adapts the template to the current statement.
\end{itemize}

\subsection{Additional Skills}

\begin{itemize}[left = 0em]
    \item \textbf{Mathematical Formula} ($o_{\mathrm{math}}$) derives a closed form for counting, combinatorics, or number-theory tasks, often giving $O(1)$ or $O(\log n)$ time.

    \item \textbf{Symmetry} ($o_{\mathrm{sym}}$) merges equivalent states to reduce the state space.

    \item \textbf{Randomization} ($o_{\mathrm{random}}$) introduces randomized algorithms when exact deterministic methods are too costly.

    \item \textbf{Approximation} ($o_{\mathrm{approx}}$) relaxes exact optimality and returns an efficient approximate solution with a stated quality target.

    \item \textbf{Caching} ($o_{\mathrm{cache}}$) adds memoization to recursive computation, turning repeated subproblems into a finite state table.

    \item \textbf{Parallelization} ($o_{\mathrm{parallel}}$) identifies independent sub-computations and schedules them concurrently.
\end{itemize}

Table~\ref{tab:skills_full} summarizes the preconditions, state effects, complexity changes, and failure modes of all skills.

\begin{table*}[t]
  \centering
  \scriptsize
  \setlength{\tabcolsep}{1pt}
  \renewcommand{\arraystretch}{1.5}
  \caption{Complete algorithmic skill library. Each skill $o_i$ is specified by its precondition $\phi_i^{\mathrm{pre}}$, effect on the design state, expected complexity change, and common failure mode.}
  \label{tab:skills_full}
  \begin{tabularx}{\textwidth}{p{3.0cm}XXXX}
    \toprule
    \hline
    \textbf{Skill} & \textbf{Precondition} & \textbf{Effect on State} & \textbf{Complexity Change} & \textbf{Key Failure Mode} \\
    \midrule
    Problem Abstraction ($o_{\mathrm{abs}}$) & Raw problem not typed & Sets problem and task type & $O(|q|)$ & Wrong task type \\
    Constraint Reading ($o_{\mathrm{constraint}}$) & Explicit constraints & Sets feasible complexity budget & $O(1)$ & Wrong complexity bound \\
    Brute Force ($o_{\mathrm{brute}}$) & Small input is feasible & Builds correct slow baseline & $T_{\mathrm{brute}}$ & TLE on large input \\
    Monotonicity Detection ($o_{\mathrm{mono}}$) & Ordered decision target & Adds binary-search reduction & $O(U)\to O(\log U)$ & Non-monotone predicate \\
    Invariant Design ($o_{\mathrm{inv}}$) & Iterative or greedy process & Adds correctness invariant & $O(T)$ & Invalid invariant \\
    State Design ($o_{\mathrm{state}}$) & Overlapping subproblems & Defines state and transition & $T_{\mathrm{DP}}$ & Missing state variable \\
    Decomposition ($o_{\mathrm{decomp}}$) & Separable structure & Splits into subproblems & $\sum_i T(q_i)+T_{\mathrm{combine}}$ & False independence \\
    Relaxation ($o_{\mathrm{relax}}$) & Hard constraints & Solves softened problem & $T_{\mathrm{relax}}+T_{\mathrm{round}}$ & Large gap after rounding \\
    Exchange Argument ($o_{\mathrm{exchange}}$) & Candidate greedy rule & Checks greedy choice & $O(T)$ & Failed exchange proof \\
    Data-Structure Sub. ($o_{\mathrm{ds}}$) & Repeated query pattern & Replaces scan with data structure & $Q\cdot O(\log n)$ & Wrong operation mapping \\
    Duality ($o_{\mathrm{dual}}$) & Known primal-dual form & Maps to dual problem & $T_{\mathrm{dual}}$ & Invalid dual mapping \\
    Counterexample ($o_{\mathrm{counter}}$) & Candidate algorithm exists & Generates adversarial tests & $T_{\mathrm{test}}$ & Incomplete coverage \\
    Complexity Refinement ($o_{\mathrm{complex}}$) & Correct but slow algorithm & Replaces bottleneck & $T_{\mathrm{new}}<T_{\mathrm{old}}$ & Wrong bottleneck \\
    Analogy ($o_{\mathrm{analogy}}$) & Template match exists & Maps to known paradigm & $T_{\mathrm{template}}$ & Surface match only \\
    Math Formula ($o_{\mathrm{math}}$) & Large $n$ or pattern & Derives closed form & $O(1)$ or $O(\log n)$ & Wrong formula \\
    Symmetry ($o_{\mathrm{sym}}$) & Symmetric states exist & Merges equivalent states & $T_{\mathrm{reduced}}$ & Invalid equivalence \\
    Randomization ($o_{\mathrm{random}}$) & Exact method too costly & Adds randomized method & $T_{\mathrm{rand}}$ & Unstable result \\
    Approximation ($o_{\mathrm{approx}}$) & Exact optimization too hard & Relaxes optimality target & $T_{\mathrm{approx}}$ & Weak approximation \\
    Caching ($o_{\mathrm{cache}}$) & Repeated subproblems & Adds memoization & Exp $\to |\mathcal{S}|$ & Missing cache key \\
    Parallelization ($o_{\mathrm{parallel}}$) & Independent subproblems & Schedules parallel computation & $T_{\mathrm{parallel}}$ & Hidden dependency \\
    \hline
    \bottomrule
  \end{tabularx}
\end{table*}

\section{Implementation Details}
\label{app:implementation_details}

Unless otherwise stated, all LLM components use GPT-4o-mini~\citep{openai2026models} as the backbone. Each method is evaluated with five sampling runs per problem. During inference, AlgoSkill uses an MCTS budget of $B=32$ rollouts per problem and exploration coefficient $\beta=1.0$. The correctness reward weights are set to $w_1=0.1$, $w_2=0.4$, $w_3=0.3$, and $w_4=0.2$ for compilation, unit tests, stress tests, and oracle agreement, respectively. The efficiency weights are set to $\lambda_T=\lambda_M=0.5$ and $\lambda_R=0.3$.

\end{document}